\documentclass{article}

\usepackage{arxiv}
\usepackage{placeins}
\usepackage[utf8]{inputenc}
\usepackage[T1]{fontenc}
\usepackage{hyperref}
\usepackage{url}
\usepackage{booktabs}
\usepackage{amsfonts}
\usepackage{nicefrac}
\usepackage{subcaption}
\usepackage{tabularx}
\usepackage{multirow}

\usepackage{geometry}
\usepackage{amsmath,amssymb}
\usepackage{graphicx}
\usepackage{enumitem}
\usepackage{verbatim}
\usepackage{adjustbox}
\graphicspath{ {./images/} }
\usepackage[most]{tcolorbox}
\usepackage{float}

\setlist{topsep=4pt, itemsep=2pt, parsep=0pt, partopsep=0pt}

\title{Language Models as Interfaces, Not Oracles: A Hybrid LLM–ML System for Pediatric Appendicitis}

\author{
  Soheyl Bateni \quad
  Maryam Abdolali\thanks{Corresponding author. \\ Emails: \texttt{s.bateni@email.kntu.ac.ir, maryam.abdolali@kntu.ac.ir}}\\[0.5em]
  K. N. Toosi University of Technology, Tehran, Iran
}
\date{}

\begin{document}
\renewcommand{\today}{}
\maketitle

\begin{abstract}
Large language models (LLMs) are increasingly applied to clinical decision support because they can interpret free-text documentation, yet their use as stand-alone diagnostic engines remains limited by sensitivity to prompt phrasing, information order, heuristic reasoning, and plausible but incorrect outputs. Conversely, supervised machine learning (ML) models trained on structured clinical data can deliver robust risk prediction but are difficult to integrate with narrative clinical workflows.
We present ClaMPAPP (\textbf{C}linical \textbf{La}nguage-assisted \textbf{M}achine-learning \textbf{P}ipeline for \textbf{App}endicitis), a hybrid architecture that reassigns the LLM from decision-maker to interface. The system parses note-like clinical narratives into a schema-constrained feature representation, applies deterministic plausibility checks to screen implausible extractions before inference, and passes validated features to an XGBoost classifier trained on established clinical, laboratory, and ultrasound variables. We evaluated ClaMPAPP in two independent pediatric cohorts from German hospitals against end-to-end LLM baselines spanning open-source and proprietary models. To enable controlled evaluation with preserved ground truth, narratives were synthesized from structured electronic health records through template rendering and constrained LLM rewriting, then assessed under both standard and sentence-order-permuted conditions designed to investigate positional robustness. ClaMPAPP achieved the strongest overall diagnostic performance in both internal and external validation while minimizing missed appendicitis cases---the outcome of greatest safety concern in acute triage. Comparator LLMs exhibited unstable sensitivity--specificity trade-offs and marked degradation under narrative reordering; by contrast, ClaMPAPP remained comparatively robust, including when the same base model was restricted to extraction rather than direct classification.
These findings support an LLM-as-interface, ML-as-predictor design that separates natural-language usability from predictive inference, offering a more auditable and safety-oriented pathway for clinical decision support than unconstrained generative diagnosis. Although demonstrated in pediatric appendicitis, this integration pattern is readily transferable to other medical diagnostic domains in which validated ML-based prediction models and narrative clinical documentation coexist.
\end{abstract}


\section{Introduction}

Acute appendicitis is among the most common surgical causes of acute abdominal pain in children \cite{abdpain_children,statpearls_ped_appendicitis}. Although appendicitis can occur at any age, its incidence is highest during adolescence, partly because lymphoid follicular hyperplasia is more common during this period. Along with fecaliths, this hyperplasia can obstruct the appendiceal lumen and contribute to the development of appendicitis~\cite{abdpain_children,statpearls_ped_appendicitis}. Prompt diagnosis is especially important in younger children, who may present with atypical or nonspecific symptoms. Delayed diagnosis can increase the risk of serious complications, including perforation, diffuse peritonitis, and sepsis~\cite{hpp_ped_appendicitis,statpearls_ped_appendicitis}.

These risks make the diagnostic process more challenging. In suspected pediatric appendicitis, ultrasound is generally recommended as the initial imaging modality owing to its safety profile and lack of ionizing radiation; however, its diagnostic performance remains dependent on operator expertise and the clinical setting~\cite{quigley2013us_appendicitis,tomography_US_variability,statpearls_ped_appendicitis}. Although computed tomography (CT) provides high diagnostic accuracy, concerns regarding radiation exposure in children have motivated risk-based diagnostic pathways aimed at reducing unnecessary CT utilization~\cite{ct_peds_risk,statpearls_ped_appendicitis}. Together, these diagnostic constraints highlight the need for decision-support approaches that can integrate heterogeneous clinical information and support faster, safer, and more consistent interpretation.

Artificial intelligence has emerged as a promising response to this need through two complementary paradigms: supervised machine learning (ML) and large language models (LLMs). Supervised ML models trained on tabular clinical and ultrasound variables have demonstrated strong diagnostic performance for pediatric appendicitis and related tasks, including management and severity prediction~\cite{regensburg_ml_appendicitis}. In standardized evaluations, these models have often outperformed traditional point-based scoring systems~\cite{regensburg_ml_appendicitis,ml_appendicitis_wjes}. However, their integration into clinical workflows remains challenging. Most structured ML predictors require predefined tabular inputs, whereas clinical information in real-world practice is often documented as narrative text. Moreover, model performance may degrade under dataset shift across institutions, patient populations, and clinical workflows, highlighting the need for external validation and robust deployment practices~\cite{finlayson2021clinician}.

This documentation gap makes LLMs attractive as interfaces for clinical free text. LLMs have shown strong performance on medical licensing-style examinations~\cite{liu2024chatgpt,brin2024gpt} and have also been evaluated on benchmark clinical tasks~\cite{lm_medicine}. However, growing evidence suggests that end-to-end LLM decision-making remains unreliable in realistic clinical settings~\cite{hager_llm_clinical}. Key vulnerabilities include sensitivity to prompt phrasing~\cite{hager_llm_clinical,lee2023benefits}, sensitivity to information order~\cite{hager_llm_clinical}, and the generation of authoritative-sounding but incorrect outputs, often referred to as hallucinations, which can create safety risks in important clinical decisions~\cite{lee2023benefits}. This creates a key challenge: structured ML models can provide effective predictions but are difficult to connect with free-text clinical narratives, whereas LLMs offer natural-language usability but are not reliable enough to serve as autonomous diagnostic agents.

To address this gap between natural-language usability and diagnostic reliability, we introduce \textbf{ClaMPAPP} (\textbf{C}linical \textbf{La}nguage-assisted \textbf{M}achine-learning \textbf{P}ipeline for \textbf{App}endicitis), a hybrid system that reassigns the LLM to an interface role rather than a decision-making role. ClaMPAPP uses an LLM for structured feature extraction and clinician-facing explanations, while delegating risk prediction to an XGBoost model~\cite{chen2016xgboost} trained on validated clinical variables; although instantiated here for appendicitis, the same architecture is disease-agnostic and can in principle connect narrative documentation to validated tabular predictors across other medical specialties. We further incorporate a deterministic feature-validation layer as a safety gate to check extracted variables before model inference. Although this component is specific to ClaMPAPP, its role is consistent with Good Machine Learning Practice principles for medical device development, including data quality assurance, transparency, validation, and post-deployment monitoring~\cite{fda2021gmlp}. To evaluate this architecture under controlled conditions while preserving ground-truth validity, we systematically synthesize narrative inputs from real patient tabular data using a standardized template and an LLM-based rewriting step to approximate natural clinical notes. By avoiding raw clinical-note processing, this design focuses on decision-support reliability rather than on potential noise introduced by clinical NLP, such as unstandardized abbreviations and misspelled words in clinical notes~\cite{kalyan2020secnlp}, as well as documentation artifacts associated with copy-and-paste~\cite{odonnell2009copy,hirschtick2006copy} and copy-forward practices~\cite{odonnell2009copy}, and it complements prior work highlighting the complexity of robust clinical feature-extraction pipelines~\cite{neha2026radiomics}.

Building on this methodological foundation, our specific contributions are as follows:
\begin{enumerate}

    \item We reposition the LLM as a \textbf{Feature Extractor} rather than a \textbf{Decision-Maker}. Instead of using the LLM directly for diagnosis or fine-tuning it for diagnostic classification, our framework constrains it to semantic parsing of clinical narratives. Risk estimation is then performed by a structured ML model, making the final prediction more deterministic, transparent, and auditable.
    \item We demonstrate \textbf{Safety-Oriented Diagnostic Performance} compared with direct LLM prompting. Across internal and external cohorts, ClaMPAPP achieves higher overall F1-scores and substantially reduces false negatives compared with standalone LLM baselines. This reduction in missed appendicitis cases is especially important in acute-care triage, where false negatives carry substantial clinical risk.
    \item We show improved \textbf{Robustness to Narrative Perturbations} through semantically invariant sentence-order permutation experiments. Whereas end-to-end LLM baselines show substantial performance degradation due to positional bias, ClaMPAPP remains comparatively stable and preserves its safety-first diagnostic profile.
\end{enumerate}

The remainder of this paper is organized as follows. Section~\ref{sec:related} reviews related work. Section~\ref{sec:methods} describes the ClaMPAPP system architecture, study cohorts, and narrative-generation procedure. Section~\ref{sec:experiments} presents the experimental setting, validation results, and robustness analyses. Section~\ref{sec:discussion} discusses clinical implications and limitations. Finally, Section~\ref{sec:conclusion} concludes the paper.

\section{Related Work}
\label{sec:related}

ClaMPAPP lies at the intersection of pediatric appendicitis risk stratification, reliable machine learning on structured clinical data, and the emerging use of LLMs in medicine. This section summarizes prior work and motivates the need for a hybrid ``LLM-as-interface, ML-as-predictor'' architecture.

\subsection{Pediatric Appendicitis Scores and ML Models}
Clinical scoring systems, including the Alvarado Score~\cite{alvarado1986practical} and the Pediatric Appendicitis Score (PAS)~\cite{samuel2002pediatric}, are widely used to support pediatric appendicitis risk stratification. These tools combine symptoms, physical examination findings, and selected laboratory results into numeric scores that can aid clinical decision-making; however, they are generally not sufficient as stand-alone diagnostic tests~\cite{di_saverio2020_wses_appendicitis}.

Despite these limitations, such scores remain clinically informative. Evaluations of pediatric appendicitis datasets have shown that Alvarado and PAS scores differ significantly between appendicitis and non-appendicitis groups and are among the more informative individual predictors, together with laboratory markers such as WBC count, neutrophil percentage, and CRP, as well as ultrasound findings such as appendix diameter~\cite{regensburg_ml_appendicitis}. 
Nevertheless, their routine clinical use may be constrained by the subjective or operator-dependent nature of some inputs, as well as by variability in the availability and reliability of required variables across clinical settings.~\cite{regensburg_ml_appendicitis}.

To address these limitations, recent work has increasingly adopted supervised ML classifiers trained on structured features. In pediatric appendicitis cohorts, tree-based ensemble methods such as random forests and gradient boosting have shown promising performance for diagnosis and, in some settings, for severity classification \cite{lam2023ai_appendicitis_review,regensburg_ml_appendicitis}.
Similarly, in pediatric cohorts with abdominal pain and suspected appendicitis, random forest and gradient-boosting models achieved high AUROC for appendicitis diagnosis in internal validation, although their performance decreased under external validation across hospitals \cite{external_validation}.

Collectively, these findings suggest that the underlying clinical variables remain valuable, but their predictive utility is maximized when modeled with non-linear ML rather than simple additive rules. A critical bottleneck remains: these models require structured tabular input, whereas real-world clinical information is frequently recorded as free text.

\subsection{LLMs in Medicine: Capabilities vs.\ Reliability}
LLMs are increasingly being explored for healthcare applications, including clinical documentation tasks such as note-taking and clinical summary generation, medical question answering, and support for clinical reasoning~(see \cite{lm_medicine} and references there in). Models such as GPT-4~\cite{nori2023capabilities} and medically oriented variants (e.g., Med-PaLM~2~\cite{singhal_medpalm2}) have achieved strong performance on medical benchmarks and USMLE-style evaluations.
Nevertheless, recent work cautions that strong performance on medical benchmarks and licensing-style examinations does not directly translate into safe clinical decision-making.\cite{hager_llm_clinical}.

In realistic clinical settings, LLM-based clinical tools may be affected by prompt formulation and limited contextual information, while clinical AI systems more broadly are vulnerable to dataset shift; these limitations can contribute to unreliable or incorrect outputs and underscore the need for clinician oversight and ongoing monitoring~\cite{kim_llm_reasoning,finlayson2021clinician}. In addition, hallucinated outputs and fabricated or false information remain important safety concerns when LLMs are used to support sensitive clinical decision-making or summarize clinical narratives.~\cite{hager_llm_clinical,clinical_summarization_stanford}. LLMs are increasingly explored for medical applications, including clinical documentation tasks such as note-taking and summary generation, as well as medical question answering~\cite{lm_medicine}.

\subsection{Hybrid Systems and the ClaMPAPP Approach}
Hybrid AI systems aim to combine complementary strengths of different components to improve reliability and usability in clinical workflows. One prominent family of approaches is Retrieval-Augmented Generation (RAG), which provides LLMs with retrieved external context to improve factual grounding and reduce hallucinations~\cite{Neha2025_RAG_Healthcare}. However, in many RAG-based systems, the LLM remains responsible for generating the final natural-language output. Although RAG can improve factual grounding by injecting external knowledge into LLMs, current systems still face key limitations: retrieved documents may be noisy or contain misinformation, and LLMs heavily depend on this content and remain prone to unreliable, hallucinated generation. In addition, even minor prompt perturbations can alter the retrieved passages and lead to factually incorrect answers, underscoring the need for systematic robustness evaluation in high-stakes applications.
~\cite{chen2023rgb,hu2024prompt_perturbation_rag}.

Rather than allowing the LLM to generate the clinical decision, ClaMPAPP restricts the LLM to semantic parsing of free-text narratives into structured clinical features. Risk is then computed by an XGBoost model trained on clinically validated variables, reducing exposure to probabilistic generation at the decision stage. This ``LLM-as-interface, ML-as-predictor" design aligns with guidance emphasizing validation, transparency, and monitoring in clinical ML systems~\cite{fda2021gmlp,lee2023benefits}.

Table~\ref{tab:comparison} summarizes the architectural differences between common paradigms and ClaMPAPP.

\begin{table}[t]
\centering
\footnotesize
\setlength{\tabcolsep}{6pt}
\renewcommand{\arraystretch}{1.1}

\caption{Architectural comparison of ClaMPAPP versus common AI paradigms for clinical decision support. ClaMPAPP minimizes hallucination risk by delegating the decision to a validated ML model and constraining the LLM to interface functions.}
\label{tab:comparison}

\begin{tabular}{lcccc}
\toprule
\textbf{Approach} & \textbf{Core Decision Maker} & \textbf{Input Modality} & \textbf{Explainability Source} & \textbf{Hallucination Risk} \\
\midrule
Pure LLM & LLM & Free text & LLM text & High \\
RAG systems & LLM (+ retrieval) & Free text & Retrieved context + LLM & Moderate \\
Traditional ML & ML model & Structured table & Feature importance/attribution & Low \\
\textbf{ClaMPAPP (ours)} & \textbf{ML model (XGBoost)} & \textbf{Free text} & \textbf{ML attribution + LLM explanation} & \textbf{Low} \\
\bottomrule
\end{tabular}

\vspace{2pt}
\end{table}
\begin{figure*}[!ht]
  \centering
  \includegraphics[width=\textwidth]{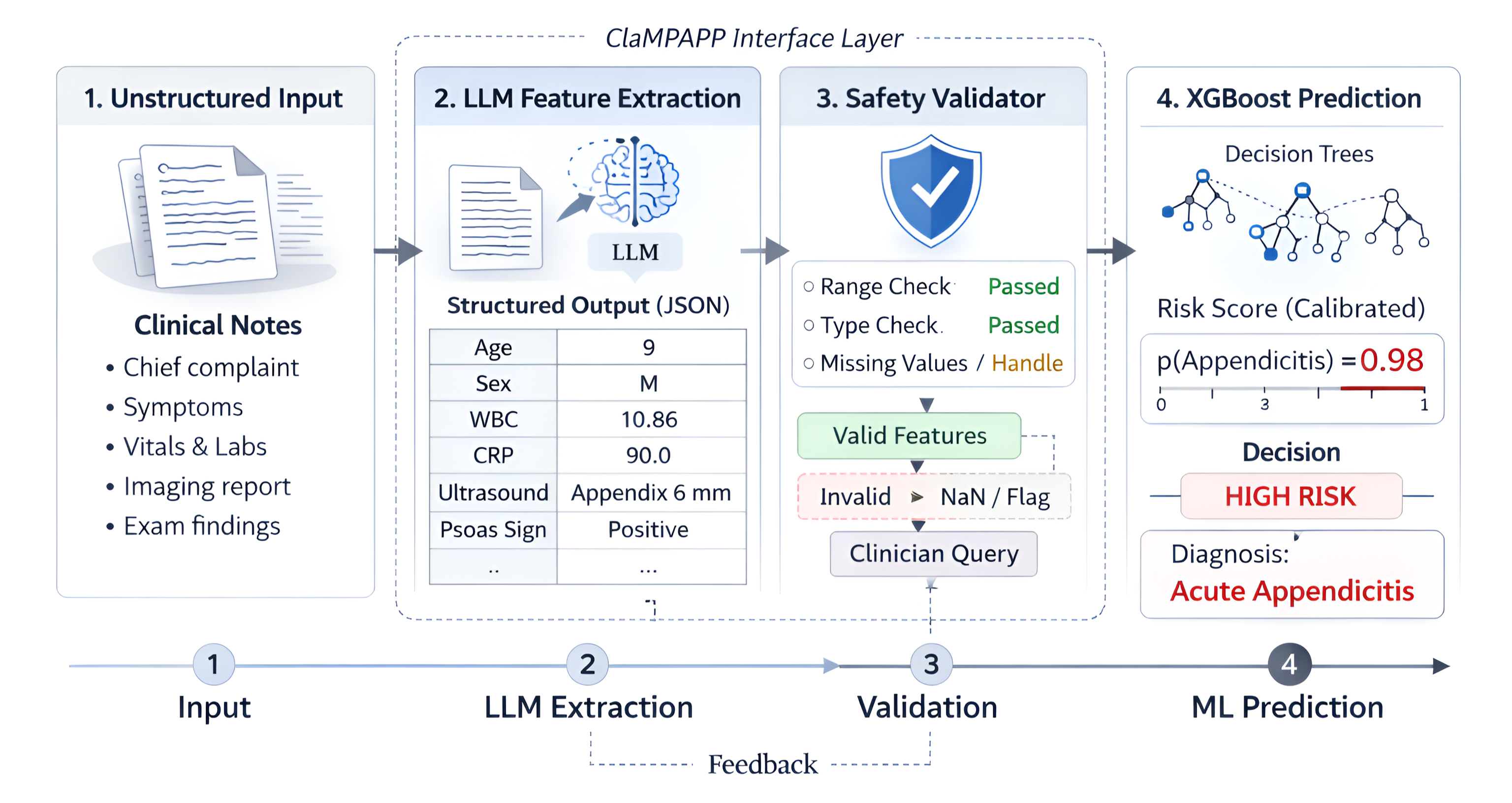}
  \caption{
    \textbf{ClaMPAPP Hybrid System Architecture.} 
    The end-to-end pipeline consists of four main stages: 
    (1) Unstructured clinical notes serve as the input. 
    (2) The ClaMPAPP Interface Layer uses an LLM for feature extraction, converting free text into a structured JSON output. 
    (3) A Safety Validator acts as a quality gate, performing range and type checks. Invalid data is either handled automatically(e.g., converted to NaN) or triggers a query back to the clinician. 
    (4) The validated feature vector is passed to the XGBoost model for a final risk prediction and diagnosis.
  }
  \label{fig:clampapp_architecture}
\end{figure*}

\section{Materials and Methods}
\label{sec:methods}

To implement the proposed hybrid approach, ClaMPAPP combines an LLM-based clinical interface with a deterministic feature validation layer and an XGBoost diagnostic classifier. The system is designed to process note-like clinical narratives, extract structured appendicitis-related features, validate these features for clinical plausibility, and generate an appendicitis risk score using a supervised tabular model. In this section, we describe the ClaMPAPP system architecture and the generation of note-like clinical narratives used for controlled evaluation. 

\FloatBarrier

\subsection{Overview of the ClaMPAPP System}
\label{subsec:clampapp_overview}

ClaMPAPP is a hybrid diagnostic decision-support pipeline for pediatric appendicitis that combines the complementary strengths ofLLMs and structured ML classifiers. The LLM component enables flexible extraction of clinical variables from natural-language case narratives, whereas the XGBoost component performs risk prediction using validated structured features.

At a high level, the ClaMPAPP workflow consists of five sequential stages:
\begin{enumerate}
    \item \textbf{Input:} A clinician or evaluator provides a free-text, note-like clinical narrative describing a pediatric patient with suspected appendicitis.
    \item \textbf{LLM-based feature extraction:} The narrative is processed by an LLM interface that maps unstructured text into a schema-constrained structured representation.
    \item \textbf{Feature validation:} A deterministic validation gate checks extracted values against predefined clinical and physiological plausibility constraints.
    \item \textbf{Risk prediction:} The validated feature vector is passed to an XGBoost classifier, which outputs an appendicitis risk score.
    \item \textbf{Report generation:} The system presents the predicted risk together with the extracted and validated clinical features in a structured clinician-facing summary.
\end{enumerate}

Figure~\ref{fig:clampapp_architecture} summarizes the end-to-end architecture of this pipeline. 

A core component of ClaMPAPP is the transformation of unstructured clinical narratives into structured diagnostic variables. This task is complicated by the heterogeneous nature of clinical notes, which often contain unstandardized abbreviations and misspellings, as well as by the limitations of conventional text representations, which are typically high-dimensional and sparse \cite{kalyan2020secnlp}. To address these challenges, ClaMPAPP uses an LLM-based extractor with task-specific tool calling to map free text into the downstream prediction schema. Specifically, the pipeline employs a dedicated \texttt{extract\_features} tool to parse the clinical narrative and generate the schema-constrained structured output required by the machine-learning classifier. The extraction schema comprises demographic and anthropometric variables, clinical scores, symptoms and physical examination findings, laboratory variables, and ultrasound-related findings. Table~\ref{tab:clampapp_feature_schema} lists the complete set of structured variables extracted from the clinical narrative.

\begin{table*}[t]
    \centering
    \caption{Structured feature schema extracted from free-text clinical narratives by the \texttt{extract\_features} tool in ClaMPAPP.}
    \label{tab:clampapp_feature_schema}
    \small
    \setlength{\tabcolsep}{6pt}
    \renewcommand{\arraystretch}{1.3}
    \begin{tabularx}{\textwidth}{>{\raggedright\arraybackslash}p{3.4cm} >{\raggedright\arraybackslash}X}
        \toprule
        \textbf{Category} & \textbf{Extracted variables} \\
        \midrule
        \textbf{Demographics and anthropometrics} &
        Age, Sex, BMI, Height, Weight \\
        \midrule
        \textbf{Clinical scores} &
        Alvarado\_Score, Pediatric\_Appendicitis\_Score \\
        \midrule
        \textbf{Symptoms and examination findings} &
        Migratory\_Pain, Lower\_Right\_Abd\_Pain, Contralateral\_Rebound\_Tenderness, Ipsilateral\_Rebound\_Tenderness, Coughing\_Pain, Nausea, Loss\_of\_Appetite, Dysuria, Stool, Peritonitis, Psoas\_Sign, Body\_Temperature \\
        \midrule
        \textbf{Laboratory variables} &
        WBC\_Count, Neutrophil\_Percentage, Neutrophilia, RBC\_Count, Hemoglobin, RDW, Thrombocyte\_Count, CRP, Ketones\_in\_Urine, RBC\_in\_Urine, WBC\_in\_Urine \\
        \midrule
        \textbf{Ultrasound and encounter-related variables} &
        US\_Performed, US\_Number, Appendix\_on\_US, Appendix\_Diameter, Free\_Fluids, Length\_of\_Stay \\
        \bottomrule
    \end{tabularx}
\end{table*}

Both the Alvarado Score and the Pediatric Appendicitis Score are calculated deterministically, and only when all of their constituent variables are available; in cases of incomplete information, these scores are not computed, thereby preventing them from being inferred from partial data.

Because a key risk in LLM-based clinical extraction is hallucination, that is, the generation of plausible-looking but incorrect values, we implemented a deterministic \emph{Feature Validator} between the LLM interface and the downstream classifier. The validator enforces physiologically and clinically plausible constraints, including bounds such as $0 \le \text{Age} \le 18$, $35.0 \le \text{Temperature} \le 42.0^{\circ}\text{C}$, and $3.0 \le \text{WBC} \le 50.0$ in units of $10^9$/L. Rather than clipping invalid values to the nearest boundary, any violated constraint is mapped to a missing value, \texttt{NaN}. This design avoids introducing biased or misleading inputs and instead converts implausible extractions into loss of information. This strategy is particularly suitable for XGBoost, which handles missing values natively by learning default split directions during training.

We employed \emph{XGBoost} as the core diagnostic engine because of its strong performance in clinical classification tasks and its suitability for structured clinical variables, including its ability to handle missing data \cite{ml_appendicitis_wjes}. The classifier was trained to predict appendicitis versus no appendicitis using structured clinical, laboratory, and ultrasound variables derived from the extracted narratives. 
 Because XGBoost handles missing values natively, missing or invalidated features were not imputed before prediction and were passed directly to the classifier. 

At inference time, the extraction--validation--prediction pipeline operates sequentially. A clinician provides a free-text narrative; the LLM parses the text into a structured output conforming to the prediction schema; the Feature Validator sanitizes the extracted values by converting implausible or out-of-range entries to \texttt{NaN}; and the XGBoost model consumes the validated feature vector to output an appendicitis risk score, interpreted as the model probability. Rather than generating an unconstrained free-form rationale, the LLM then acts as a clinical summarizer by contextualizing the XGBoost risk score alongside the extracted and validated features in a structured clinician-facing report that may include non-prescriptive triage language, such as ``High risk: consider surgical consultation'', while avoiding definitive diagnostic or treatment directives. Because the XGBoost component outputs a continuous appendicitis risk score, downstream evaluation considered not only threshold-based classification performance but also the reliability of the predicted probabilities as absolute risk estimates; these analyses are reported in the section~\ref{sec:experiments}.

\subsection{Study Cohorts}
\label{subsec:study_cohorts_methods}

This study used two independent retrospective cohorts of pediatric patients evaluated for suspected appendicitis, both originally curated as structured electronic health record (EHR) datasets.
The first cohort was derived from a retrospective series of children admitted with abdominal pain to the Department of Pediatric Surgery at St.\ Hedwig Hospital, Regensburg, Germany \cite{marcinkevics2024interpretable}. Data were extracted from the EHRs of 782 hospitalized patients. For each case, the dataset included demographic variables, clinical presentation variables, laboratory measurements, ultrasound findings, clinical scores, and the recorded management strategy.

The second cohort was obtained from a retrospective dataset of 301 children and adolescents aged 0--17 years who presented with abdominal pain and suspected appendicitis to the Department of Pediatric Surgery and Traumatology at Florence Nightingale Hospital, D{\"u}sseldorf, Germany \cite{external_validation}. Patients with prior appendectomy, chronic intestinal disease, or ongoing antibiotic treatment at the time of admission were excluded. This dataset contained a comparable set of demographic, clinical, laboratory, and ultrasound variables, enabling independent external evaluation.

Both cohorts were released as structured tabular records rather than free-text clinical notes. These tabular variables served as the source data for the narrative-generation procedure described in Section~\ref{subsec:narrative_construction}.

\subsection{Generation of Note-Like Clinical Narratives}
\label{subsec:narrative_construction}

\begin{figure*}[!ht]
  \centering
  \includegraphics[width=\textwidth]{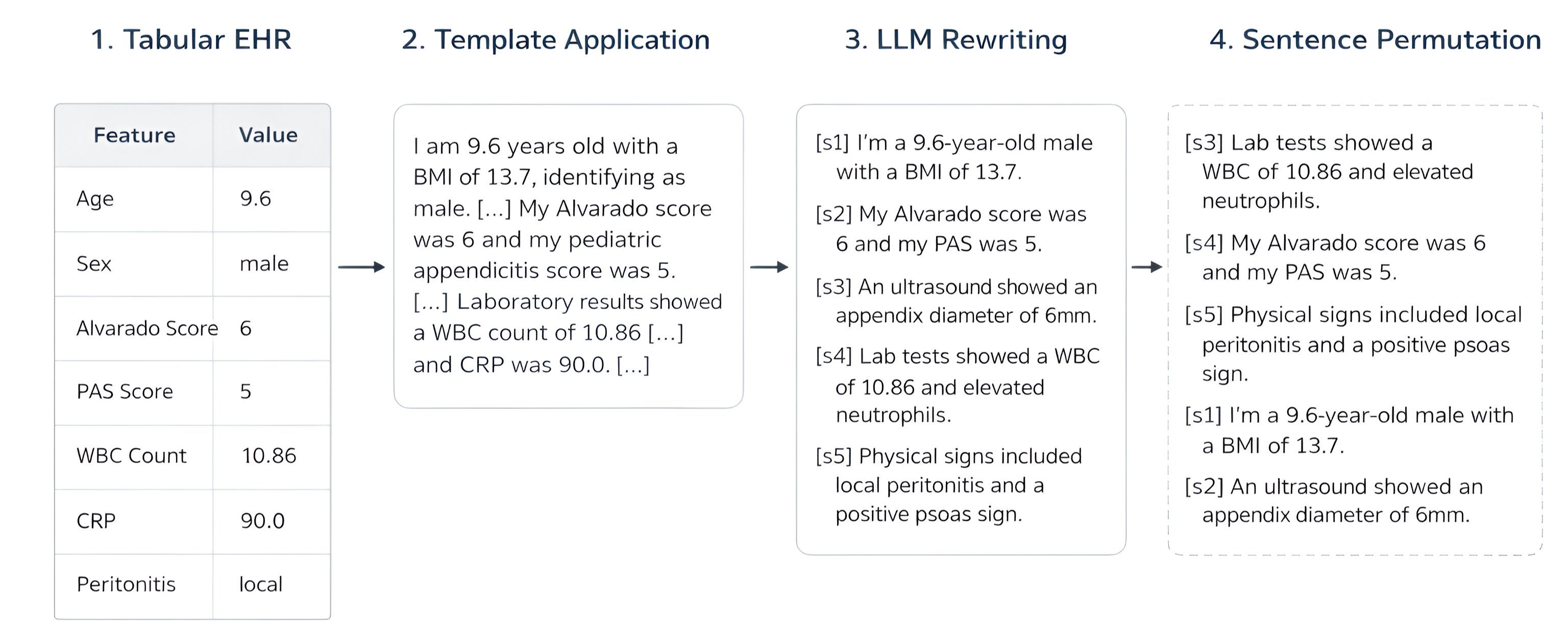}
  \caption{
    \textbf{Data and Narrative Generation Pipeline.}
    This figure illustrates the four-stage process used to generate note-like clinical narratives from structured EHR variables.
    \textbf{(1) Tabular EHR:} the pipeline begins with raw structured patient data.
    \textbf{(2) Template Application:} tabular fields populate a predefined template, producing a standardized narrative.
    \textbf{(3) LLM Rewriting:} an LLM, Llama-3.1-8B, rewrites the template into more natural first-person text under strict preservation constraints.
    \textbf{(4) Sentence Permutation:} to evaluate robustness to ordering effects while preserving coherence, sentence order is modified using a bipartite structural inversion, implemented as a swap between the first and second halves of the narrative. The resulting text is then used as input to the ClaMPAPP system.
  }
  \label{fig:narrative_generation}
\end{figure*}

Although clinical documentation is often available as free text, both cohorts used in this study were released as structured tabular variables extracted from electronic health records. Previous work has used natural language generation to generate textual summaries of patient histories~\cite{scott2013_datatotext} 
and to generate synthetic clinical text from structured EHR variables~\cite{lee2018_nlg_ehr}, 
as well as to convert tabular EHR into pseudo-notes compatible with large language models~\cite{lee2025_pseudonotes}.
 Related synthetic benchmark work has also paired structured clinical variables with generated unstructured notes for evaluating clinical information extraction~\cite{rabaey2025simsum}. To enable a controlled evaluation of the LLM-based extraction pipeline, we reconstructed narrative inputs from real-world tabular records using a standardized data-to-text procedure.
Importantly, all values contained in the generated narratives correspond to measured or documented variables from real patient encounters; only the format was transformed from table to text, while the underlying clinical content remained unchanged.
The narrative-generation pipeline comprised four stages: processing raw structured patient data, template-based rendering, LLM-based rewriting, and controlled two-part template permutation. In the third stage, the template-based clinical narrative generated from the structured tables was rewritten by an LLM to increase variation in the generated narratives and make them more closely resemble reports written by humans. In the fourth stage, the template-based clinical narrative was divided into two predefined textual segments, and their order was swapped. The resulting permuted template was then also rewritten by the LLM to produce an alternative narrative variant for model evaluation. Fidelity to the original data was assessed separately at the end of the pipeline through manual sampling and auditing. An overview of this process is provided in Figure ~\ref{fig:narrative_generation}.

To create textual inputs for the LLM component, we defined a fixed template that converts each patient's tabular record into a structured English case narrative. The template was written in the first person to approximate note-like clinical documentation and encoded all available variables with standardized wording, ordering, and units. Curly braces denote placeholders populated with patient-specific values:

\begin{tcolorbox}[
  enhanced,
  colback=gray!5,
  colframe=gray!60!black,
  title={Patient Case Narrative Template (Excerpt)},
  fontupper=\small\ttfamily,
  arc=0mm, boxrule=0.5pt
]
I am \{Age\} years old with a BMI of \{BMI\}, identifying as \{Sex\}. 
I stand \{Height\} cm tall and weigh \{Weight\} kg.

My Alvarado score was \{Alvarado\_Score\} and my pediatric appendicitis score was \{Pediatric\_Appendicitis\_Score\}.

On ultrasound, the appendix was noted as \{Appendix\_on\_US\}, measuring \{Appendix\_Diameter\} mm in diameter.

I reported migratory pain: \{Migratory\_Pain\}, as well as right lower quadrant pain: \{Lower\_Right\_Abd\_Pain\}, contralateral rebound tenderness: \{Contralateral\_Rebound\_Tenderness\}, and pain upon coughing: \{Coughing\_Pain\}.

I experienced nausea: \{Nausea\} and loss of appetite: \{Loss\_of\_Appetite\}. 
My temperature was \{Body\_Temperature\}~$^\circ$C.

\ldots
\end{tcolorbox}

For brevity, only an excerpt of the template is shown above; the complete template and generation code will be made publicly available in the accompanying repository upon publication. For each patient in the Regensburg and D{\"u}sseldorf cohorts, the corresponding tabular values were inserted into this template.
Because the templated narratives were complete but stylistically rigid, we used the Llama-3.1-8B model as a constrained rewriting model to produce more natural note-like text while preserving the underlying ground-truth content. The model was instructed to rewrite each structured template into fluent first-person English while strictly preserving all clinical facts, numerical values, and binary findings, without adding interpretation, diagnosis, or commentary. The rewriting prompt was specified as follows:

\begin{tcolorbox}[
  enhanced,
  sharp corners,
  boxrule=0.8pt,
  colback=gray!5,
  colframe=gray!60!black,
  title={Narrative Rewriting Prompt},
  fontupper=\small\ttfamily
]
You are a professional medical-text editor. Your only job is to rewrite 
the user's English text about a case related to appendicitis. Follow these rules exactly:
\begin{enumerate}[leftmargin=*]
    \item ONLY rewrite the provided text to make it sound natural and human-written.
    \item PRESERVE every clinical feature, sign, symptom, and factual detail. 
          Do NOT remove, change, or add any clinical facts.
    \item Do NOT provide explanations, diagnoses, or commentary.
    \item Return EXACTLY one rewritten text block and NOTHING else.
    \item If the original text contains contradictions, keep the original medical content 
          but rewrite the phrasing to be natural.
\end{enumerate}
\end{tcolorbox}

This stage transformed rigid placeholder-based expressions into more natural note-like language and thereby provided a more realistic testbed for evaluating downstream feature extraction. An illustrative excerpt of the rewritten output is shown below:

\begin{tcolorbox}[
  enhanced,
  sharp corners,
  boxrule=0.8pt,
  colback=green!5, 
  colframe=green!40!black,
  title={Excerpt of an LLM-Rewritten Case Narrative},
  fontupper=\small
]
I'm 12 years and 8 months old, with a body mass index of 16.9. I'm female and 148 centimeters tall, weighing 37 kilograms. \ldots I experienced pain in my lower right abdomen, which I described as a migratory pain, and had tenderness on the opposite side of my body when I pressed. \ldots Laboratory tests showed 7.7 white blood cells per microliter, and my CRP level was zero. \ldots An ultrasound of my abdomen revealed that my appendix measured 7.1 millimeters in diameter and showed signs of being inflamed. \ldots
\end{tcolorbox}

To assess robustness to variation in information ordering, we generated permuted narratives using a bipartite structural inversion. Each narrative was divided into two sequential halves, and the order of these halves was then reversed. This procedure altered the position of clinical information within the note while preserving overall coherence and avoiding the degradation associated with fully random sentence shuffling. The same procedure was applied to both the Regensburg and D{\"u}sseldorf cohorts.

Because LLM outputs may be sensitive to instruction phrasing, information order, and information quantity, and may not reliably follow instructions
~\cite{hager_llm_clinical}, we performed a manual fidelity audit to compare the rewritten narratives against the original tabular ground truth. A random sample of $31$ clinical reports was drawn from the Regensburg evaluation split, corresponding to $15.35\%$ of the $N=202$ internal evaluation cohort. For each report, we verified the exact preservation of approximately $20$ core clinical features, including demographics, vital signs, laboratory values, physical examination findings, imaging findings, and clinical scores.
At the report level, the $31$ sampled cases demonstrated high fidelity: $14$ reports showed perfect fidelity, $13$ contained exactly one feature-level deviation, and $4$ contained exactly two feature-level deviations. No report contained more than two deviations. Across the audited sample, corresponding to approximately $620$ feature-level checks, we identified $21$ isolated feature-level errors, equivalent to $3.4\%$. These deviations comprised feature omissions ($n=10$), contradictions ($n=7$), qualitative abstractions ($n=3$), and one null imputation ($n=1$), in which a missing value, \texttt{NaN}, was rendered as zero in the generated text.
To assess generalizability across datasets, we repeated the audit on a random sample of $10$ reports from the D{\"u}sseldorf cohort. This audit covered $17$ features, corresponding to approximately $170$ feature-level checks. The results were similar: $7$ reports showed perfect fidelity, $1$ report contained exactly one deviation, and $2$ reports contained exactly two deviations. At the feature level, we observed five deviations: three omissions, one contradiction, and one qualitative abstraction, in which a tabular ``no'' was paraphrased as ``normal''.

Overall, the fidelity audits indicated that the rewriting stage introduced natural linguistic variation while maintaining a low rate of content drift. With feature-level error rates of $3.4\%$ for Regensburg and $2.9\%$ for D{\"u}sseldorf, the pipeline preserved more than $96\%$ of the audited information across both cohorts, supporting the use of the rewritten narratives as a high-fidelity proxy for downstream feature-extraction experiments.

\section{Experiments}
\label{sec:experiments}
In this section, we present a comprehensive evaluation of the ClaMPAPP system using two independent retrospective cohorts of pediatric patients evaluated for suspected appendicitis. We first introduce the experimental setting, including cohort characteristics, input representations, model training procedures, and evaluation metrics. We then identify the optimal machine learning backbone by analyzing the baseline performance of multiple candidate models. Next, we assess the comparative diagnostic performance of ClaMPAPP relative to baseline approaches and clinically relevant reference standards. Finally, we investigate the robustness of the system by examining its sensitivity to input permutation.

\subsection{Experimental Setting}
\label{subsec:experimental_setting}

The two study cohorts were used in distinct roles for model development and evaluation. The Regensburg cohort served as the internal dataset for training and in-domain evaluation, whereas the D{\"u}sseldorf cohort served exclusively as an external validation dataset for assessing out-of-site generalization.

\begin{figure*}[!ht]
  \centering
  \includegraphics[width=0.7\textwidth]{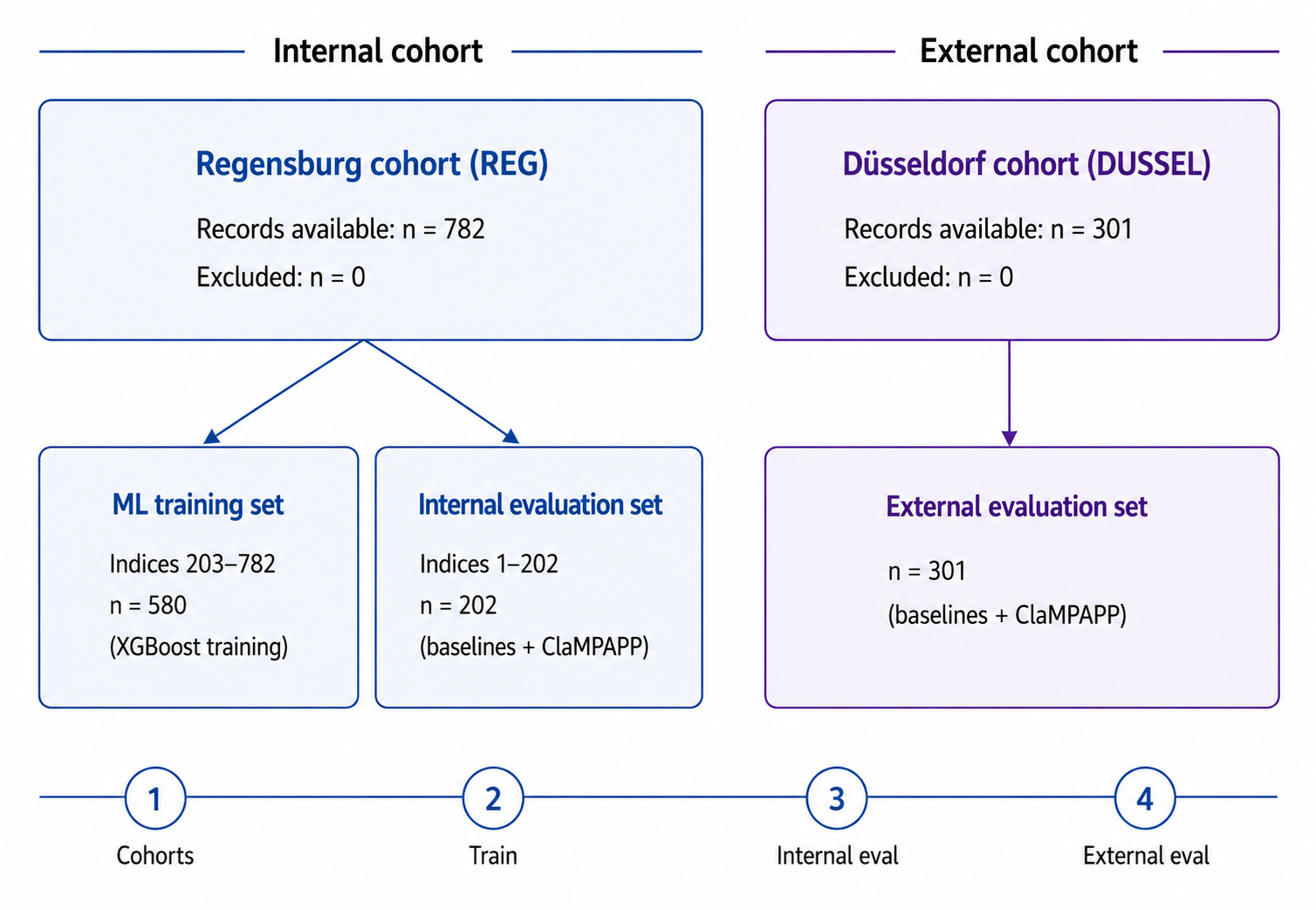}
  \caption{
    \textbf{Cohort Flow Diagram (STROBE-style).} 
    This diagram details the selection and allocation of patient records for the ClaMPAPP study. 
    The \textbf{internal cohort (Regensburg)} was split into a training set ($n=580$) for the XGBoost model and an internal evaluation set ($n=202$) for performance assessment. 
    The \textbf{external cohort (D{\"u}sseldorf)}, with $n=301$ records, was used as a distinct set for external validation to test the model's generalization capabilities.
  }
  \label{fig:cohort_flow}
\end{figure*}

As shown in Figure~\ref{fig:cohort_flow}, the Regensburg cohort was divided into a training set of $n=580$ patients for XGBoost model development and an internal evaluation set of $n=202$ patients for performance assessment. The D{\"u}sseldorf cohort, comprising $n=301$ records, was not used for model training, prompt refinement, or threshold selection, and was reserved entirely for external evaluation.

This experimental design enabled assessment of both internal performance and generalizability across institutions using independent retrospective cohorts.

\paragraph{Baseline Models and Performance Metrics}

We benchmarked ClaMPAPP against a spectrum of LLMs spanning general-purpose models, medical-domain architectures, and state-of-the-art proprietary models to probe different diagnostic behaviors:
\begin{enumerate}
    \item \textbf{medgemma-4b-it}~\cite{sellergren2025_medgemma}: a medical-domain instruction-tuned model derived from the Gemma family.
    \item \textbf{llama-3.1-8b-instant}~\cite{grattafiori2024_llama3}: a general-purpose instruction-tuned model.
    \item \textbf{OpenAI GPT-5.5}~\cite{openai2026_gpt55_system_card}: a state-of-the-art proprietary large language model.
    \item \textbf{OpenAI o3-pro}~\cite{openai2025_o3_o4mini_system_card}: an advanced proprietary model optimized for complex reasoning.
    \item \textbf{Anthropic Claude Opus 4.7}~\cite{anthropic2026_claude_opus47_system_card}: a highly capable proprietary model known for instruction adherence.
    \item \textbf{oasst-sft-1-pythia-12b}~\cite{kopf2023openassistant}: an open-source conversational assistant model.
\end{enumerate}




Diagnostic performance was quantified using standard classification metrics: accuracy; sensitivity (recall) which is critical for minimizing missed diagnoses; specificity, important for reducing unnecessary interventions; precision; and F1-score.



\subsection{Baseline Performance: Establishing the Optimal ML Backbone}
To establish a robust classification engine for our hybrid system, we benchmarked several supervised machine learning algorithms. All models were trained on the Regensburg cohort and evaluated on both the internal hold-out test set and the external D{\"u}sseldorf cohort to assess generalization capabilities. Detailed performance metrics were recorded for both internal (Table~\ref{tab:test1_internal}) and external (Table~\ref{tab:test2_external}) validation sets.

\begin{table}[htbp]
  \centering
  \caption{Performance metrics on the internal validation set (TEST 1).}
  \label{tab:test1_internal}
  \resizebox{\textwidth}{!}{%
    \begin{tabular}{lccccccccc}
    \toprule
    \textbf{Model Name} & \textbf{Acc} & \textbf{Sen} & \textbf{Spec} & \textbf{Prec} & \textbf{F1} & \textbf{TP} & \textbf{TN} & \textbf{FP} & \textbf{FN} \\
    \midrule
    LightGBM & $0.815$ & $0.988$ & $0.687$ & $0.700$ & $0.820$ & $84$ & $79$ & $36$ & $1$ \\
    \textbf{XGBoost} & \textbf{0.895} & \textbf{0.988} & \textbf{0.826} & \textbf{0.808} & \textbf{0.889} & \textbf{84} & \textbf{95} & \textbf{20} & \textbf{1} \\
    Decision Tree & $0.870$ & $0.906$ & $0.843$ & $0.811$ & $0.856$ & $77$ & $97$ & $18$ & $8$ \\
    Random Forest & $0.890$ & $0.824$ & $0.939$ & $0.909$ & $0.864$ & $70$ & $108$ & $7$ & $15$ \\
    Logistic Regression & $0.785$ & $0.871$ & $0.722$ & $0.698$ & $0.775$ & $74$ & $83$ & $32$ & $11$ \\
    \bottomrule
    \end{tabular}%
  }
\end{table}

\begin{table}[htbp]
  \centering
  \caption{Performance metrics on the external validation set (TEST 2).}
  \label{tab:test2_external}
  \resizebox{\textwidth}{!}{%
    \begin{tabular}{lccccccccc}
    \toprule
    \textbf{Model Name} & \textbf{Acc} & \textbf{Sen} & \textbf{Spec} & \textbf{Prec} & \textbf{F1} & \textbf{TP} & \textbf{TN} & \textbf{FP} & \textbf{FN} \\
    \midrule
    LightGBM & $0.807$ & $0.935$ & $0.394$ & $0.833$ & $0.881$ & $215$ & $28$ & $43$ & $15$ \\
    \textbf{XGBoost} & \textbf{0.811} & \textbf{0.917} & \textbf{0.465} & \textbf{0.847} & \textbf{0.881} & \textbf{211} & \textbf{33} & \textbf{38} & \textbf{19} \\
    Decision Tree & $0.738$ & $0.817$ & $0.479$ & $0.836$ & $0.826$ & $188$ & $34$ & $37$ & $42$ \\
    Random Forest & $0.767$ & $0.874$ & $0.423$ & $0.831$ & $0.852$ & $201$ & $30$ & $41$ & $29$ \\
    Logistic Regression & $0.781$ & $0.939$ & $0.268$ & $0.806$ & $0.867$ & $216$ & $19$ & $52$ & $14$ \\
    \bottomrule
    \end{tabular}%
  }
\end{table}

\FloatBarrier

Tables~\ref{tab:test1_internal} and~\ref{tab:test2_external} show a clear decrease in performance when models were evaluated on the external cohort, consistent with reduced transportability under cohort shift. Although Random Forest achieved the highest internal specificity ($0.939$), its external specificity decreased substantially to $0.423$, accompanied by a drop in accuracy from $0.890$ to $0.767$. The Decision Tree model showed a similar pattern, with accuracy decreasing from $0.870$ to $0.738$ and specificity decreasing from $0.843$ to $0.479$. These results suggest weaker external generalization for the single-tree and bagging-based models.


We selected the Gradient Boosting framework, implemented as XGBoost, as the ML backbone for ClaMPAPP because it provided a favorable balance between internal performance and external generalization. XGBoost achieved the highest internal Accuracy ($0.895$) and F1 score ($0.889$), while sharing the highest internal Sensitivity ($0.988$) with LightGBM. On the external cohort, it achieved the highest Accuracy ($0.811$), the highest Precision ($0.847$), and a joint-best F1 score ($0.881$, tied with LightGBM), indicating comparatively stable performance under cohort shift. Although its external Specificity decreased to $0.465$, XGBoost maintained high Sensitivity ($0.917$) and Precision ($0.847$), supporting reliable identification of the target class while acknowledging reduced specificity in the external setting.

\subsection{Comparative Diagnostic Performance}
\label{sec:llm_results}
\begin{figure}[htbp]
    \centering
    \begin{subfigure}[b]{0.48\textwidth}
        \centering
        \includegraphics[width=\textwidth]{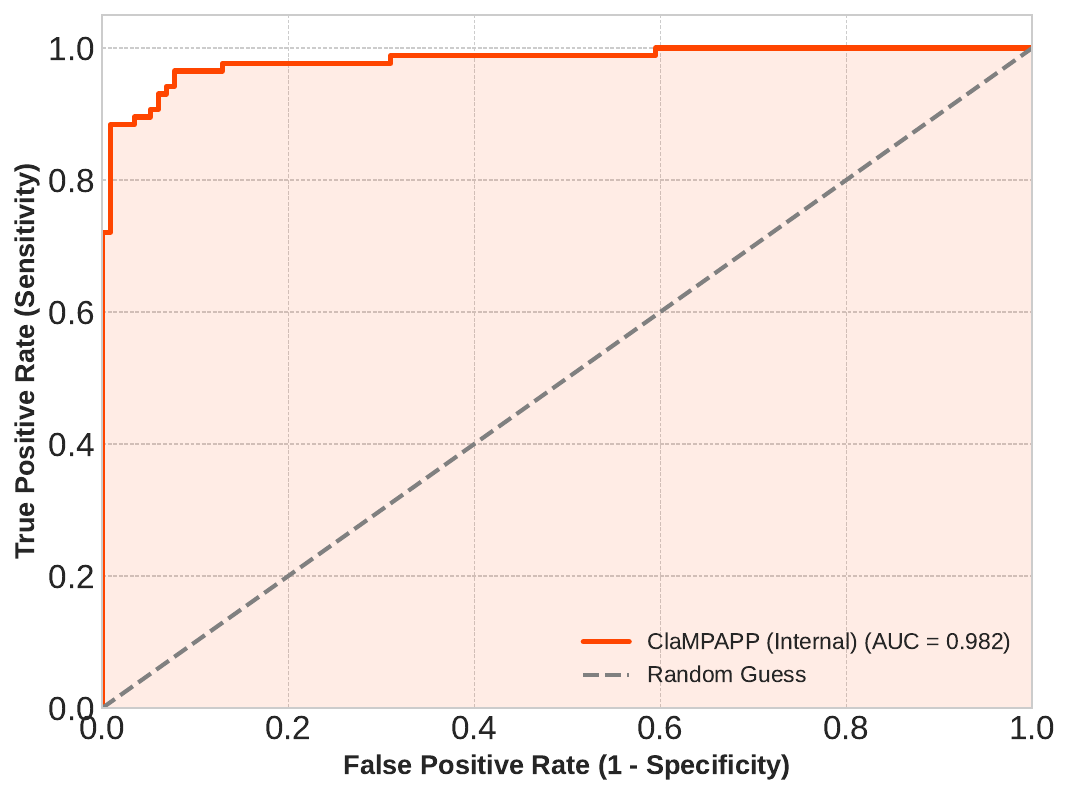}
        \caption{ROC Curve (AUC = 0.982)}
        \label{fig:roc_internal}
    \end{subfigure}%
    \hfill
    \begin{subfigure}[b]{0.48\textwidth}
        \centering
        \includegraphics[width=\textwidth]{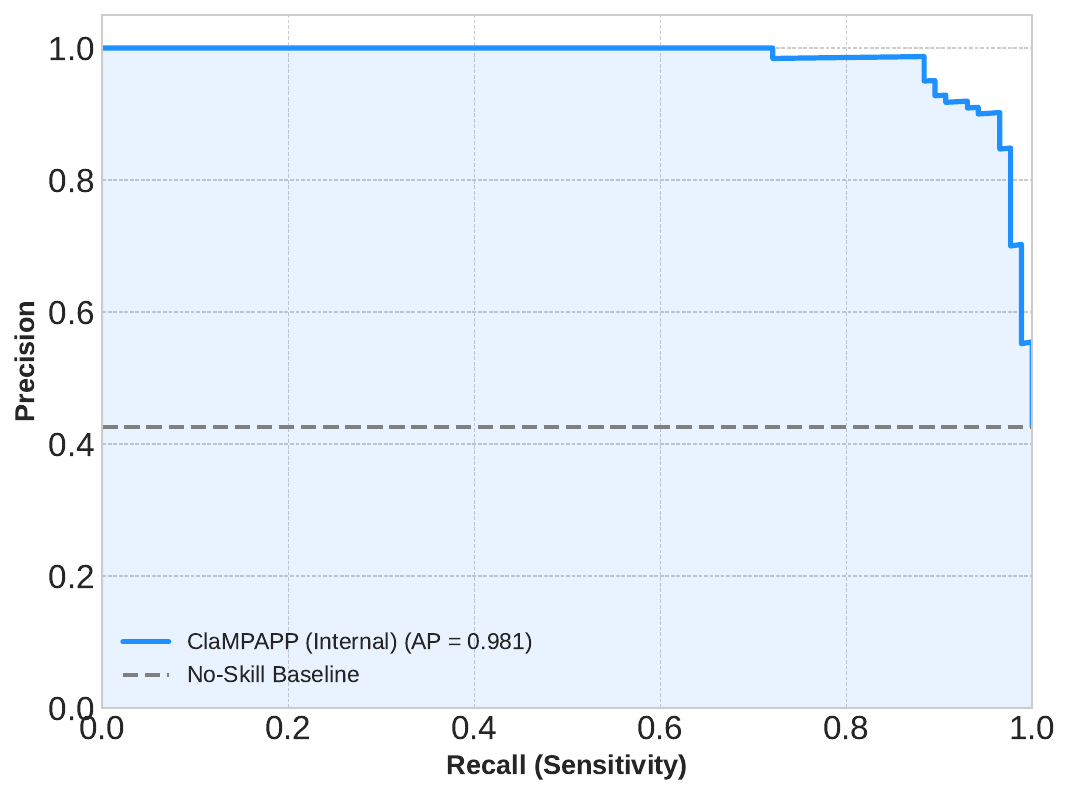}
        \caption{Precision-Recall Curve (AP = 0.981)}
        \label{fig:pr_internal}
    \end{subfigure}
    
    \vspace{0.5em}
    \caption{
        \textbf{Performance analysis of the ClaMPAPP system on the internal validation cohort.} 
        (\subref{fig:roc_internal}) The ROC curve indicates an AUC of 0.982, demonstrating excellent discrimination. 
        (\subref{fig:pr_internal}) The Precision-Recall curve shows an Average Precision (AP) of 0.981. 
        The high average precision indicates strong overall ranking performance and suggests that the lower point-estimate precision reported in Table~\ref{tab:internal_results} reflects the trade-off of discrete classification rather than limited model discrimination.
    }
    \label{fig:performance_curves_internal}
\end{figure}

\begin{figure}[htbp]
    \centering
    \begin{subfigure}[b]{0.48\textwidth}
        \centering
        \includegraphics[width=\textwidth]{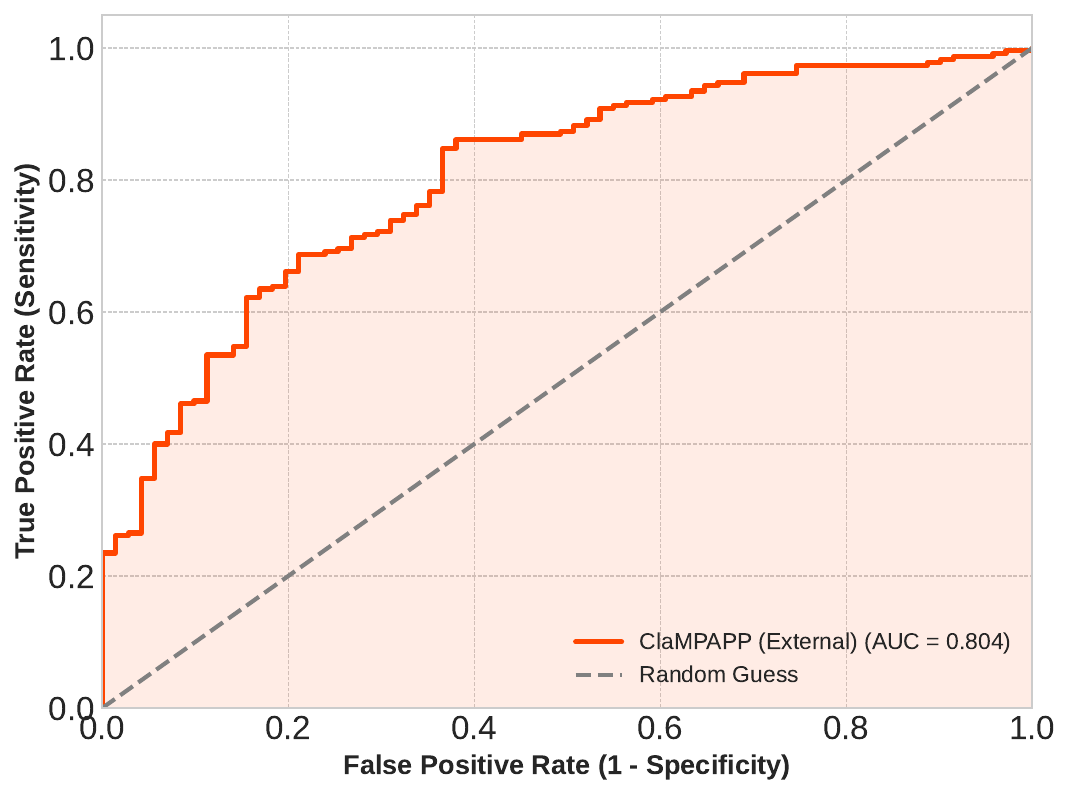}
        \caption{ROC curve (AUC $=0.804$)}
        \label{fig:roc_external}
    \end{subfigure}\hfill
    \begin{subfigure}[b]{0.48\textwidth}
        \centering
        \includegraphics[width=\textwidth]{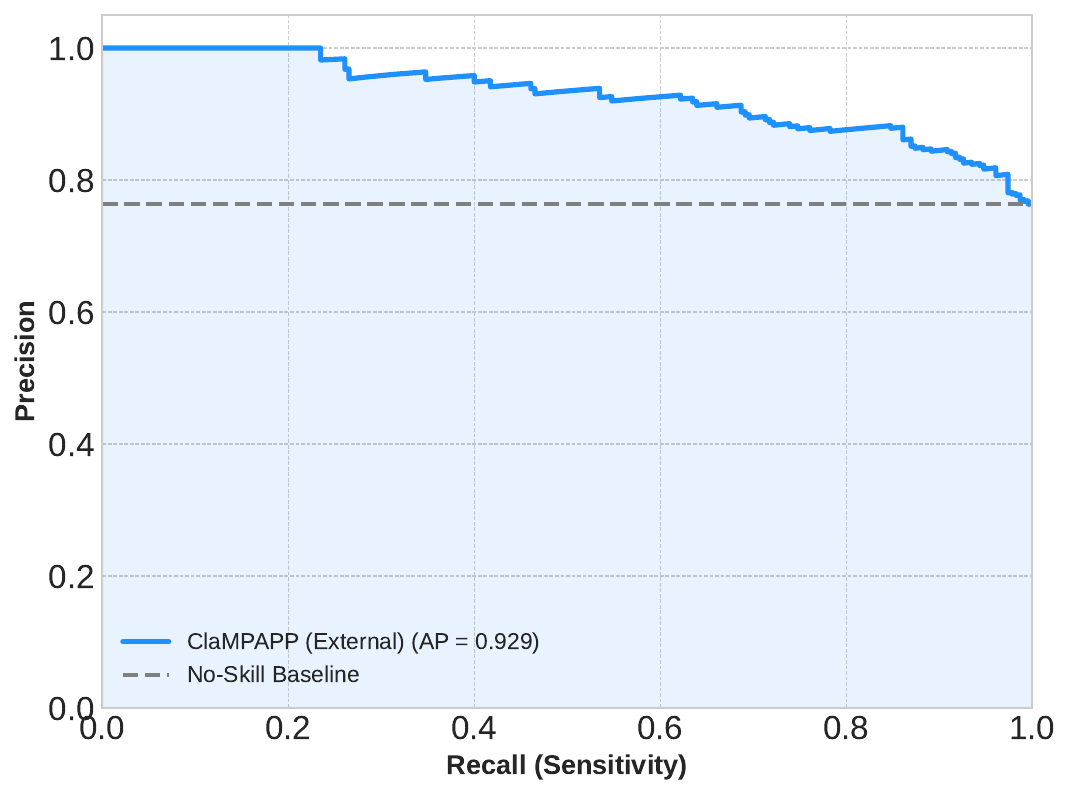}
        \caption{Precision--Recall curve (AP $=0.929$)}
        \label{fig:pr_external}
    \end{subfigure}
    
    \caption{
        \textbf{External validation metrics on the D{\"u}sseldorf cohort.}
        (\subref{fig:roc_external}) The ROC curve shows an AUC of $0.804$, indicating generalization to an independent clinical site.
        (\subref{fig:pr_external}) The Precision--Recall curve maintains a high Average Precision (AP) of $0.929$, indicating strong ranking performance despite distribution shift.
    }
    \label{fig:performance_curves_external}
\end{figure}

We evaluated the end-to-end hybrid system, ClaMPAPP, against a panel of representative baseline LLMs, including both open-source and closed-source models, on the internal and external cohorts. Because the baseline LLMs were constrained to produce discrete binary predictions using a strict JSON output format, they did not provide calibrated or continuous risk scores. We therefore report threshold-independent metrics, including ROC and Precision--Recall (PR) curves (Figures~\ref{fig:performance_curves_internal} and~\ref{fig:performance_curves_external}), only for ClaMPAPP, which outputs continuous risk scores from the XGBoost classifier.

On the internal Regensburg cohort (Table~\ref{tab:internal_results}; Figure~\ref{fig:performance_curves_internal}), ClaMPAPP provided the best overall discrimination among the evaluated models, achieving the highest accuracy ($85.1\%$) and F1 score ($84.8\%$) while maintaining high sensitivity ($97.7\%$). Although its specificity ($75.9\%$) was not the highest overall, it exceeded that of most LLM baselines and avoided the extreme false-positive rate observed for the highest-sensitivity baseline, OASST-Pythia-12b. ClaMPAPP missed only $2$ confirmed appendicitis cases ($\mathrm{FN}=2$), supporting a safety-oriented classification profile with a more favorable sensitivity--specificity trade-off than the LLM-only models.


\begin{table}[htbp]
\centering
\caption{
Performance comparison on the internal Regensburg cohort. Bold indicates the best value in each column (maximum for Acc/Sen/Spec/Prec/F1/TP/TN, and minimum for FP/FN). Although some LLM baselines achieve higher scores in individual metrics (e.g., specificity for MedGemma), ClaMPAPP offers the most favorable overall clinical trade-off, achieving the highest Accuracy and F1 score while maintaining very low false-negative counts.}
\label{tab:internal_results}
\resizebox{\textwidth}{!}{%
\begin{tabular}{lccccccccc}
\toprule
\textbf{Model} & \textbf{Acc} & \textbf{Sen} & \textbf{Spec} & \textbf{Prec} & \textbf{F1} & \textbf{TP} & \textbf{TN} & \textbf{FP} & \textbf{FN} \\
\midrule
\textbf{ClaMPAPP} & \textbf{0.851} & 0.977 & 0.759 & 0.750 & \textbf{0.848} & 84 & 88 & 28 & 2 \\
OpenAI o3-pro & 0.752 & 0.919 & 0.629 & 0.648 & 0.760 & 79 & 73 & 43 & 7 \\
OpenAI GPT-5.5 & 0.748 & 0.802 & 0.707 & 0.669 & 0.730 & 69 & 82 & 34 & 17 \\
Anthropic Claude Opus 4.7 & 0.693 & 0.581 & 0.776 & 0.658 & 0.617 & 50 & 90 & 26 & 36 \\
MedGemma-4b-it & 0.738 & 0.535 & \textbf{0.888} & \textbf{0.780} & 0.634 & 46 & 103 & 13 & 40 \\
Llama-3.1-8b-instant & 0.663 & 0.756 & 0.595 & 0.580 & 0.657 & 65 & 69 & 47 & 21 \\
OASST-Pythia-12b & 0.436 & \textbf{0.988} & 0.026 & 0.429 & 0.599 & 85 & 3 & 113 & 1 \\
\bottomrule
\end{tabular}
}
\end{table}

\FloatBarrier

Baseline LLMs, including proprietary models, demonstrated clinically relevant trade-offs. \textit{MedGemma-4b-it} and \textit{Anthropic Claude Opus 4.7} achieved high specificity ($88.8\%$ and $77.6\%$, respectively), but at the cost of substantially lower sensitivity and high false-negative counts (FN $= 40$ and $36$, respectively), which is undesirable in acute triage settings. Conversely, \textit{OASST-Pythia-12b} achieved near-perfect sensitivity ($98.8\%$) largely by predicting appendicitis for almost all patients, resulting in extremely low specificity ($2.6\%$) and a very high false-positive burden (FP $= 113$). \textit{OpenAI GPT-5.5} showed a relatively balanced baseline profile (Accuracy $= 0.748$; Specificity $= 70.7\%$), but still missed many more confirmed appendicitis cases than ClaMPAPP (FN $= 17$ vs.\ $2$). In contrast, ClaMPAPP achieved the highest overall accuracy ($85.1\%$) and F1 score ($84.8\%$), while maintaining very high sensitivity ($97.7\%$) and specificity ($75.9\%$) that was higher than most, but not all, LLM baselines.
End-to-end LLM baselines also raise a practical deployment concern: they may not always return outputs in a fixed, machine-readable format. This matters in urgent triage, where each patient requires a valid decision. By using the LLM only for validated feature extraction and generating the final score with a gradient-boosting classifier, ClaMPAPP produced a valid risk score for every patient in our evaluation.

To assess robustness under distribution shift, the models were evaluated on the independent external D{\"u}sseldorf cohort. As shown in Table~\ref{tab:external_results} and Figure~\ref{fig:performance_curves_external}, ClaMPAPP achieved the highest accuracy ($80.7\%$) and F1-score ($0.881$), indicating superior out-of-site generalization compared to baseline LLMs.

\begin{table}[htbp]
\centering
\caption{
Performance comparison on the external D{\"u}sseldorf cohort. Bold indicates the best value in each column (maximum for Acc/Sen/Spec/Prec/F1/TP/TN, and minimum for FP/FN). Although some LLM baselines achieve higher scores in individual metrics (e.g., specificity and precision for OpenAI GPT-5.5), ClaMPAPP offers the most favorable overall clinical trade-off, achieving the highest F1 score and minimal false-negative counts at a safety-first operating point.
}
\label{tab:external_results}
\resizebox{\textwidth}{!}{%
\begin{tabular}{lccccccccc}
\toprule
\textbf{Model} & \textbf{Acc} & \textbf{Sen} & \textbf{Spec} & \textbf{Prec} & \textbf{F1} & \textbf{TP} & \textbf{TN} & \textbf{FP} & \textbf{FN} \\
\midrule
\textbf{ClaMPAPP} & \textbf{0.807} & \textbf{0.935} & 0.394 & 0.833 & \textbf{0.881} & \textbf{215} & 28 & 43 & \textbf{15} \\
OpenAI o3-pro & 0.724 & 0.709 & 0.775 & 0.911 & 0.797 & 163 & 55 & 16 & 67 \\
OpenAI GPT-5.5 & 0.708 & 0.661 & \textbf{0.859} & \textbf{0.938} & 0.776 & 152 & \textbf{61} & \textbf{10} & 78 \\
Anthropic Claude Opus 4.7 & 0.704 & 0.670 & 0.817 & 0.922 & 0.776 & 154 & 58 & 13 & 76 \\
MedGemma-4b-it & 0.648 & 0.613 & 0.761 & 0.892 & 0.727 & 141 & 54 & 17 & 89 \\
Llama-3.1-8b-instant & 0.708 & 0.752 & 0.563 & 0.848 & 0.797 & 173 & 40 & 31 & 57 \\
OASST-Pythia-12b & 0.731 & 0.922 & 0.113 & 0.771 & 0.840 & 212 & 8 & 63 & 18 \\
\bottomrule
\end{tabular}
}
\end{table}

\FloatBarrier

Crucially, the external evaluation highlights the sensitivity-oriented operating profile of ClaMPAPP. Although its specificity on the D{\"u}sseldorf cohort was modest ($39.4\%$), ClaMPAPP achieved the highest sensitivity ($93.5\%$) and missed the fewest positive cases ($\mathrm{FN}=15$) among the evaluated models. This contrasts with models such as \textit{OpenAI GPT-5.5} and \textit{MedGemma-4b-it}, which achieved higher specificities ($85.9\%$ and $76.1\%$, respectively) and high precision, but missed substantially more positive cases ($\mathrm{FN}=78$ and $\mathrm{FN}=89$, respectively). Similarly, \textit{Llama-3.1-8b-instant} missed $57$ positive cases ($\mathrm{FN}=57$). Thus, ClaMPAPP appears to prioritize minimizing missed positive cases over specificity in the external cohort.


A particularly noteworthy observation across both internal and external evaluations is the substantial performance gain achieved by the hybrid architecture relative to its underlying LLM when used as an end-to-end diagnostic model. When \textit{Llama-3.1-8b} was evaluated as a standalone model, it showed clinically undesirable false-negative rates, with $21$ false negatives internally and $57$ externally. In contrast, when the same frozen Llama-3.1 model was constrained to \emph{feature extraction} and the final risk prediction was delegated to the supervised XGBoost classifier with deterministic validation, ClaMPAPP substantially reduced missed positive cases, lowering false negatives to $2$ internally and $15$ externally. This improvement was accompanied by higher overall diagnostic performance, including improved accuracy and F1 score in both cohorts. These results suggest that the hybrid ``LLM-as-interface, ML-as-predictor'' design can leverage the clinical utility of a base LLM without resource-intensive fine-tuning, providing a computationally efficient path toward deployable clinical decision support, particularly in settings where minimizing missed diagnoses is prioritized.

\subsection{Probability Calibration of ClaMPAPP}
\label{sec:calibration_results}

Because ClaMPAPP outputs continuous appendicitis risk scores, we evaluated not only discrimination but also probability calibration on both cohorts. On the internal Regensburg cohort, ClaMPAPP showed excellent discrimination (AUC-ROC $=0.982$), but the raw predicted probabilities were not perfectly calibrated. The raw Brier score was $0.118$, log loss was $0.375$, and the expected calibration error (ECE; 10 bins) was $0.167$. The calibration intercept and slope were $-2.936$ and $1.113$, respectively, and the Hosmer--Lemeshow test indicated lack of perfect fit ($p<0.001$). Thus, although the model ranked cases very effectively, the raw probability scale showed imperfect agreement with observed event frequencies.

\begin{figure*}[htbp]
    \centering

    \begin{subfigure}[t]{0.45\textwidth}
        \centering
        \includegraphics[width=\linewidth]{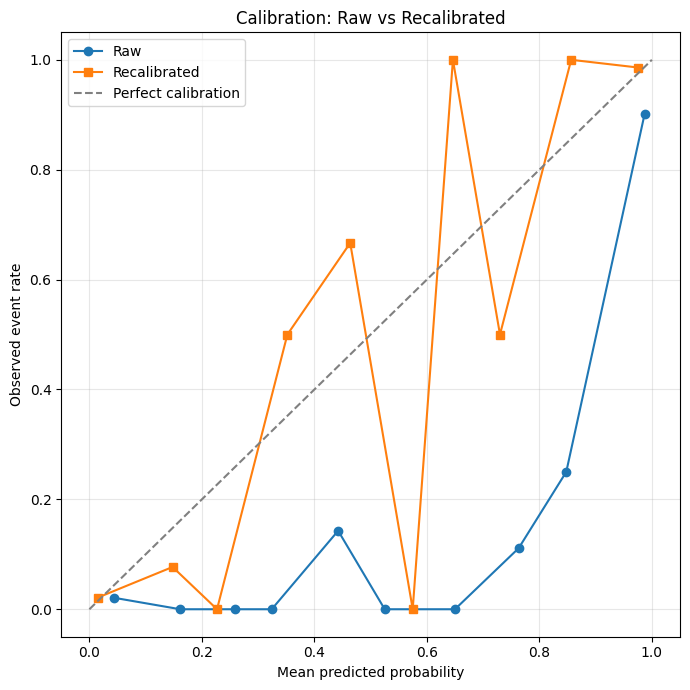}
        \caption{Regensburg cohort}
        \label{fig:calibration_regensburg}
    \end{subfigure}
    \hfill
    \begin{subfigure}[t]{0.45\textwidth}
        \centering
        \includegraphics[width=\linewidth]{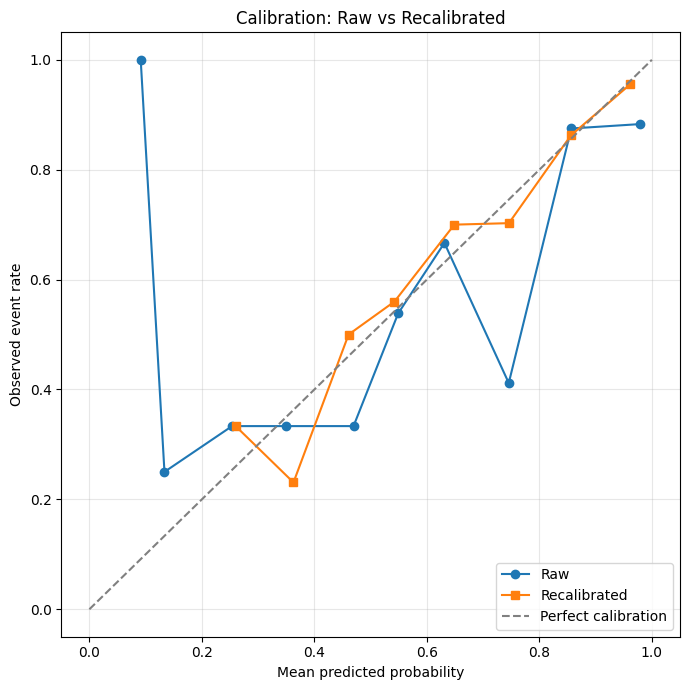}
        \caption{D{\"u}sseldorf cohort}
        \label{fig:calibration_duesseldorf}
    \end{subfigure}

    \caption{
        \textbf{Calibration of ClaMPAPP on the internal and external cohorts.}
        Reliability diagrams compare observed event rates with mean predicted probabilities across bins.
        \textbf{(A) Regensburg cohort:} raw probabilities show marked deviation from the ideal diagonal despite excellent discrimination; post hoc logistic recalibration improves apparent agreement between predicted and observed risk.
        \textbf{(B) D{\"u}sseldorf cohort:} raw probabilities also show imperfect calibration, particularly in the spread of predicted risks; recalibration again improves apparent calibration.
        In both cohorts, recalibration leaves ranking performance unchanged and primarily adjusts the interpretation of scores as absolute risk probabilities.
    }
    \label{fig:calibration_both}
\end{figure*}

On the external D{\"u}sseldorf cohort, raw calibration was also imperfect. The AUC-ROC was $0.804$, the Brier score was $0.150$, log loss was $0.508$, and ECE was $0.096$. The calibration intercept was close to zero ($-0.060$), whereas the calibration slope was substantially below $1$ ($0.513$), indicating that the spread of predicted probabilities was misaligned with observed outcome frequencies. The Hosmer--Lemeshow test again suggested lack of perfect fit ($p<0.001$). Reliability diagrams for both cohorts are shown in Figure~\ref{fig:calibration_both}.

\begin{table}[htbp]
\centering
\small
\caption{
Calibration metrics for ClaMPAPP on the internal Regensburg and external D{\"u}sseldorf cohorts. Recalibrated values are from exploratory post hoc logistic recalibration performed and evaluated on the same cohort; therefore, these results reflect \emph{apparent} calibration improvement.
}
\label{tab:calibration_results}
\begin{tabular}{lccc}
\toprule
\textbf{Cohort} & \textbf{Metric} & \textbf{Raw} & \textbf{Recalibrated} \\
\midrule
\multirow{6}{*}{Regensburg}
& ROC AUC & 0.982 & 0.982 \\
& Brier score & 0.118 & 0.046 \\
& Log loss & 0.375 & 0.164 \\
& ECE (10 bins) & 0.167 & 0.039 \\
& Calibration intercept & -2.936 & $\approx 0$ \\
& Calibration slope & 1.113 & $\approx 1$ \\
\midrule
\multirow{6}{*}{D{\"u}sseldorf}
& ROC AUC & 0.804 & 0.804 \\
& Brier score & 0.150 & 0.140 \\
& Log loss & 0.508 & 0.431 \\
& ECE (10 bins) & 0.096 & 0.026 \\
& Calibration intercept & -0.060 & $\approx 0$ \\
& Calibration slope & 0.513 & $\approx 1$ \\
\bottomrule
\end{tabular}
\end{table}

In an exploratory post hoc analysis, logistic recalibration substantially improved \emph{apparent} calibration metrics on both cohorts (Table~\ref{tab:calibration_results}). On the internal Regensburg cohort, ClaMPAPP showed excellent discrimination (AUC-ROC $=0.982$), but the raw predicted probabilities were not perfectly calibrated. The raw Brier score was $0.118$, log loss was $0.375$, and the expected calibration error (ECE; 10 bins) was $0.167$. The calibration intercept and slope were $-2.936$ and $1.113$, respectively, and the Hosmer--Lemeshow test indicated lack of perfect fit ($p<0.001$). Thus, although the model ranked cases very effectively, the raw probability scale showed imperfect agreement with observed event frequencies.

These findings suggest that ClaMPAPP’s miscalibration may be due, at least in part, to a mismatch in the probability scale rather than poor ranking ability, and that this mismatch can be improved by simple logistic recalibration. Because recalibration and evaluation were performed on the same cohorts, these results reflect apparent rather than externally validated calibration improvement.

\FloatBarrier

\subsection{Robustness Analysis: Sensitivity to Input Permutation}
\label{sec:robustness}

A known vulnerability of LLM-based systems is sensitivity to the position of relevant information within the input context: changing the position of key information can significantly alter performance.~\cite{hager_llm_clinical}. To assess robustness to this effect, we conducted a controlled sentence-order perturbation test while preserving the underlying clinical facts. Specifically, each narrative was divided into two contiguous halves, and the order of the two halves was swapped. This two-block perturbation changes where clinical information appears within the note while avoiding the incoherence introduced by fully random sentence shuffling. The resulting narratives therefore provide a fact-preserving test of whether model predictions are stable under realistic variation in clinical documentation order.

Results on the permuted Regensburg cohort are presented in Table~\ref{tab:robustness_internal}. Under sentence-order permutation, the pure LLM baselines showed generally weak and variable performance. Several models produced accuracies close to 0.5, indicating limited robustness to this perturbation. For example, \textit{Llama-3.1-8b} achieved an accuracy of 0.475 on the permuted cohort. Likewise, \textit{MedGemma-4b-it} reached a sensitivity of only 0.128, corresponding to a large number of missed appendicitis cases. Among the pure LLM baselines, \textit{OpenAI GPT-5.5} achieved the strongest overall performance on this permuted evaluation, with an accuracy of 0.624, but still produced 19 false negatives (FN$=19$). Other proprietary models, including \textit{Anthropic Claude Opus 4.7} and \textit{OpenAI o3-pro}, also showed substantial error rates, with both false positives and false negatives remaining prominent. Overall, these findings indicate that end-to-end LLM predictions in this setting remain sensitive to changes in prompt structure, despite preservation of the underlying clinical content.

In contrast, ClaMPAPP achieved the strongest overall performance on the permuted cohort and outperformed all pure LLM baselines across the principal classification metrics. Notably, it had perfect sensitivity ($1.000$) with no missed positive cases (FN$=0$), while also achieving the highest accuracy and F1 score in this evaluation. At the same time, the model is not entirely unaffected by input perturbations, and small metric differences may still arise when variation in narrative structure influences the upstream LLM-based feature extraction step. Because these extracted variables are subsequently provided to the XGBoost classifier, limited variability at the extraction stage can propagate to the final prediction. Nevertheless, the hybrid design constrains the role of the LLM to structured clinical parsing and delegates final risk estimation to a supervised tabular model, thereby reducing the impact of prompt-surface variation and preserving a more robust diagnostic profile.

\begin{table}[htbp]
\centering
\caption{
Robustness analysis on the Regensburg cohort under sentence-order permutation (\emph{bipartite structural inversion}). \textbf{Bold} indicates the best value in each column.}
\label{tab:robustness_internal}
\resizebox{\textwidth}{!}{%
\begin{tabular}{lccccccccc}
\toprule
\textbf{Model} & \textbf{Acc} & \textbf{Sen} & \textbf{Spec} & \textbf{Prec} & \textbf{F1} & \textbf{TP} & \textbf{TN} & \textbf{FP} & \textbf{FN} \\
\midrule
\textbf{ClaMPAPP} & \textbf{0.807} & \textbf{1.000} & 0.664 & \textbf{0.688} & \textbf{0.815} & \textbf{86} & 77 & 39 & \textbf{0} \\
OpenAI o3-pro & 0.515 & 0.779 & 0.319 & 0.459 & 0.578 & 67 & 37 & 79 & 19 \\
OpenAI GPT-5.5 & 0.624 & 0.779 & 0.509 & 0.540 & 0.638 & 67 & 59 & 57 & 19 \\
Anthropic Claude Opus 4.7 & 0.584 & 0.640 & 0.543 & 0.509 & 0.567 & 55 & 63 & 53 & 31 \\
MedGemma-4b-it & 0.515 & 0.128 & \textbf{0.802} & 0.324 & 0.183 & 11 & \textbf{93} & \textbf{23} & 75 \\
Llama-3.1-8b-instant & 0.475 & 0.523 & 0.440 & 0.409 & 0.459 & 45 & 51 & 65 & 41 \\
OASST-Pythia-12b & 0.446 & 0.953 & 0.069 & 0.432 & 0.594 & 82 & 8 & 108 & 4 \\
\bottomrule
\end{tabular}
}
\end{table}

\FloatBarrier

\begin{figure*}[t]
    \centering
    \includegraphics[width=\textwidth]{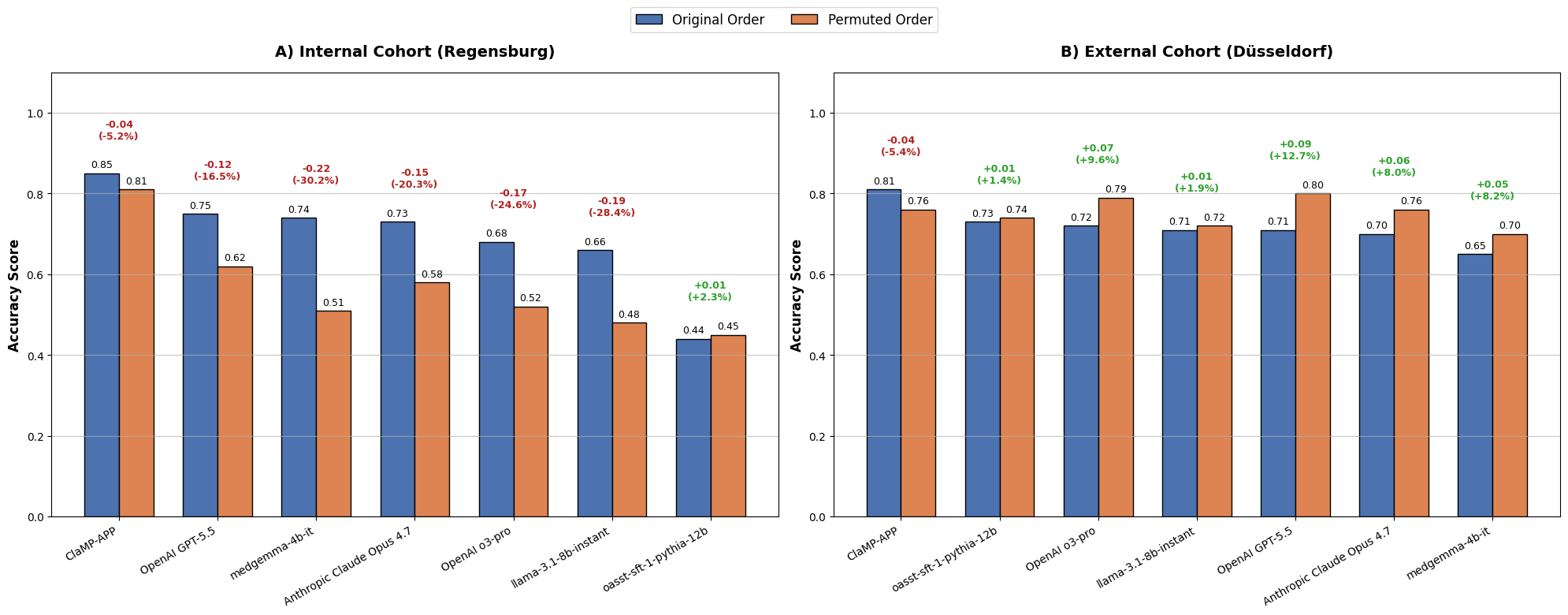}
    \caption{
        \textbf{Robustness Analysis: Accuracy Stability under Sentence-Order Permutation (Bipartite Structural Inversion).}
        Blue bars represent the original structured input, while orange bars show results after sentence-order permutation.
        \textbf{(A) Internal Cohort (Regensburg):} Baseline LLMs exhibit a significant performance collapse; for instance, MedGemma-4b-it and Llama-3.1-8b show relative accuracy drops of $30.2\%$ and $28.4\%$, respectively. In contrast, ClaMPAPP demonstrates high architectural robustness with only a $5.2\%$ decrease.
        \textbf{(B) External Cohort (D{\"u}sseldorf):} While the external dataset shows less severe degradation for baseline models, ClaMPAPP maintains the most stable and safety-oriented profile. Absolute and percentage changes are annotated above each model pair.
    }
    \label{fig:robustness_analysis}
\end{figure*}

The permutation test on the external D{\"u}sseldorf cohort (Table~\ref{tab:robustness_external}) provided additional evidence for the robustness of the hybrid architecture under distribution shift. Under sentence-order permutation, ClaMPAPP achieved the highest sensitivity (0.926) and the lowest number of missed appendicitis cases (FN$=17$), indicating the most conservative and triage-oriented operating point among all evaluated models. 

By contrast, frontier models such as \textit{OpenAI GPT-5.5} and \textit{OpenAI o3-pro} achieved higher overall accuracy (0.797 and 0.794, respectively) and markedly higher specificity (0.718 and 0.634, respectively), but at the cost of lower sensitivity (0.822 and 0.843) and substantially more missed positive cases (FN$=41$ and FN$=36$). A similar pattern was observed for other baselines, including \textit{Llama-3.1-8b-instant} (FN$=44$) and \textit{MedGemma-4b-it} (FN$=67$), which failed to detect a larger number of appendicitis cases.

Figure~\ref{fig:robustness_analysis} summarizes the accuracy changes induced by sentence-order permutation across both cohorts and shows that several baseline LLMs experienced larger performance degradation, whereas ClaMPAPP exhibited a comparatively smaller shift. Overall, these results suggest that decoupling feature extraction from decision-making helps preserve a high-sensitivity diagnostic profile under perturbation and external distribution shift. While pure LLMs may retain stronger specificity and overall accuracy, ClaMPAPP more reliably maintains the sensitivity level required for safe triage.

\begin{table}[htbp]
\centering
\caption{
Robustness analysis on the D{\"u}sseldorf cohort under sentence-order permutation (\emph{bipartite structural inversion}). \textbf{Bold} indicates the best value in each column.
}
\label{tab:robustness_external}
\resizebox{\textwidth}{!}{%
\begin{tabular}{lccccccccc}
\toprule
\textbf{Model} & \textbf{Acc} & \textbf{Sen} & \textbf{Spec} & \textbf{Prec} & \textbf{F1} & \textbf{TP} & \textbf{TN} & \textbf{FP} & \textbf{FN} \\
\midrule
\textbf{ClaMPAPP} & 0.764 & \textbf{0.926} & 0.239 & 0.798 & 0.857 & \textbf{213} & 17 & 54 & \textbf{17} \\
OpenAI o3-pro & 0.794 & 0.843 & 0.634 & 0.882 & \textbf{0.862} & 194 & 45 & 26 & 36 \\
OpenAI GPT-5.5 & \textbf{0.797} & 0.822 & \textbf{0.718} & \textbf{0.904} & 0.861 & 189 & \textbf{51} & \textbf{20} & 41 \\
Anthropic Claude Opus 4.7 & 0.761 & 0.809 & 0.606 & 0.869 & 0.838 & 186 & 43 & 28 & 44 \\
MedGemma-4b-it & 0.701 & 0.709 & 0.676 & 0.876 & 0.784 & 163 & 48 & 23 & 67 \\
Llama-3.1-8b-instant & 0.721 & 0.809 & 0.437 & 0.823 & 0.816 & 186 & 31 & 40 & 44 \\
OASST-Pythia-12b & 0.741 & 0.878 & 0.296 & 0.802 & 0.838 & 202 & 21 & 50 & 28 \\
\bottomrule
\end{tabular}
}
\end{table}

\FloatBarrier

\section{Discussion}
\label{sec:discussion}

End-to-end LLM baselines, despite their linguistic fluency and accessibility, showed variable performance on the internal Regensburg cohort and did not maintain consistently stable performance on the external Dusseldorf cohort. They were also sensitive to prompt perturbations, occasionally produced ungrounded rationales, and were influenced by information ordering (positional bias)~\cite{wornow2023shaky}. These findings align with prior evidence that strong benchmark performance does not necessarily translate into safe real-world clinical decision-making~\cite{hager_llm_clinical} and underscore the limited evidence current evaluation regimes provide for real-world utility, reliability, and safety. End-to-end LLMs are therefore unsuitable as stand-alone diagnostic tools for pediatric appendicitis in safety-critical settings.

By contrast, an XGBoost model trained on structured clinical and sonographic variables achieved strong discrimination and showed more robust behavior across cohorts, consistent with established supervised ML approaches in this domain~\cite{regensburg_ml_appendicitis,external_validation}. At the same time, discrimination did not necessarily imply perfectly calibrated probabilities: ClaMPAPP ranked cases well in both cohorts, but its raw probability estimates were not fully calibrated, particularly in the internal cohort. Exploratory post hoc logistic recalibration improved apparent calibration without changing discrimination, suggesting that at least part of the mismatch arose from the probability scale rather than from a fundamentally inadequate predictor. Conventional ML nonetheless faces practical adoption barriers, including rigid tabular inputs, limited natural-language interaction, and perceived opacity with restricted explanatory value. ClaMPAPP addresses these complementary limitations through a hybrid architecture that explicitly separates interface from decision-making, using the ML component for risk estimation from validated structured variables and the LLM for structured data extraction, interaction with unstructured clinical input, and clinician-facing reporting.

By delegating the final prediction to supervised ML and constraining the LLM to extraction and reporting, the system limits probabilistic generation at the decision stage and supports a more auditable workflow aligned with supervised, workflow-integrated LLM use rather than autonomous diagnosis~\cite{hager_llm_clinical,maity_llm_healthcare_review}. Although this study focuses specifically on pediatric appendicitis, the underlying ClaMPAPP design idea may also be explored in future work for other diseases and clinical tasks where validated structured predictors are available.

\subsection{Clinical Implications}
ClaMPAPP's defining clinical strength is high sensitivity (\textbf{$97.7\%$} internally and \textbf{$93.5\%$} externally), which is consequential for appendicitis triage given the morbidity of missed diagnosis. Lower external specificity (\textbf{$39.4\%$}) implies a higher false-positive burden, but this must be weighed against the low false-negative rate and the model's ability to identify most confirmed cases. The system should therefore be interpreted as a triage-oriented risk-stratification aid rather than a definitive diagnostic test; flagged patients still require standard clinical assessment, imaging, laboratory evaluation, and local care pathways.

Calibration analyses further indicate that outputs are more reliably interpreted as relative risk scores than as transportable absolute probabilities---acceptable for sensitivity-oriented triage, where rank ordering and threshold behavior dominate, but increasingly important for probability-based counseling, escalation, or resource allocation. The natural-language interface and optional follow-up questioning aim to reduce usability barriers in busy emergency settings, complemented by deterministic validation that prevents implausible extracted values from entering the predictor. Throughout, ClaMPAPP is intended strictly as decision support augmenting---not replacing---clinical judgment, physical examination, serial reassessment, imaging, laboratory trends, and institutional workflows.

\subsection{Limitations}

This study has several limitations. The evaluation was retrospective and therefore requires prospective validation under real-time clinical conditions. Experiments used note-like narratives synthesized from structured EHR variables rather than raw clinical notes; although this preserved ground truth for controlled benchmarking, it likely overestimates extraction robustness relative to native documentation with abbreviations, typos, copy-forward artifacts, irregular missingness, and site-specific conventions.

Both cohorts were derived from German institutions. Thus, although the hybrid architecture is broadly transferable, the underlying XGBoost classifier may require local validation, recalibration, threshold tuning, or retraining before deployment elsewhere to address dataset shift in disease incidence, measurement protocols, clinical workflows, and practice patterns~\cite{finlayson2021clinician}. Cross-lingual deployment likewise requires verification of extraction fidelity under local documentation styles, prompts, abbreviations, and terminology.

Despite the \emph{Feature Validator}, extraction may still fail in ambiguous, contradictory, or incomplete narratives, and safeguards are more effective at detecting implausible values than clinically plausible but incorrect ones. In addition, although the model outputs continuous risk scores and initial calibration analyses were performed, we did not conduct an independent prospective calibration study or assess subgroup-specific calibration. Recalibration analyses were also post hoc and evaluated on the same cohorts used to estimate recalibration parameters, reflecting apparent rather than transportable calibrated performance. Absolute predicted probabilities should therefore be interpreted cautiously until validation and recalibration are performed in the target population. 



\section{Conclusion}
\label{sec:conclusion}
We introduced \emph{ClaMPAPP}, a hybrid clinical decision-support system that combines an LLM-based interface with a supervised XGBoost classifier for pediatric appendicitis risk stratification. Using retrospective cohorts from two German hospitals, we demonstrated the feasibility of converting note-like clinical narratives derived from structured EHR variables into a validated feature schema for structured ML-based prediction. Our results suggest that end-to-end LLM baselines, despite strong language capabilities, did not provide the reliability required for safety-oriented appendicitis decision support in this study setting, whereas supervised models trained on structured clinical and sonographic variables showed stronger discriminative performance and more stable behavior under external validation. ClaMPAPP achieved high sensitivity and fewer missed appendicitis cases than the evaluated LLM baselines, supporting its role as a triage-oriented risk-stratification tool rather than a stand-alone diagnostic system. Its central contribution is the separation of language understanding from final risk estimation: the LLM is constrained to feature extraction, data completion, and clinician-facing communication, while final prediction is delegated to a validated ML model with feature validation before inference. This ``LLM-as-interface, ML-as-predictor'' paradigm may provide a practical pathway for translating validated tabular risk models into clinically usable decision-support tools across medical specialties. Future work should evaluate ClaMPAPP prospectively in real emergency-department workflows using native clinical notes and should extend the framework to broader pediatric abdominal-pain presentations and other diseases where validated structured risk models already exist.

\bibliographystyle{unsrt}
\bibliography{ref}

@article{abdpain_children,
  author  = {Reust, Carin E. and Williams, Amy},
  title   = {Acute Abdominal Pain in Children},
  journal = {American Family Physician},
  year    = {2016},
  volume  = {93},
  number  = {10},
  pages   = {830--837},
  url     = {https://www.aafp.org/pubs/afp/issues/2016/0515/p830.html}
}

@misc{statpearls_ped_appendicitis,
  author       = {Waseem, Muhammad and Wang, Cecily F.},
  title        = {Pediatric Appendicitis},
  year         = {2025},
  howpublished = {In: StatPearls [Internet]. Treasure Island (FL): StatPearls Publishing},
  note         = {Last updated June 17, 2025; accessed June 14, 2026},
  url          = {https://www.ncbi.nlm.nih.gov/books/NBK441864/}
}

@article{hpp_ped_appendicitis,
  author  = {Almaramhy, Hamdi Hameed},
  title   = {Acute Appendicitis in Young Children Less than 5 Years: Review Article},
  journal = {Italian Journal of Pediatrics},
  year    = {2017},
  volume  = {43},
  number  = {1},
  pages   = {15},
  doi     = {10.1186/s13052-017-0335-2}
}

@article{quigley2013us_appendicitis,
  author  = {Quigley, Alan J. and Stafrace, Samuel},
  title   = {Ultrasound assessment of acute appendicitis in paediatric patients: methodology and pictorial overview of findings seen},
  journal = {Insights into Imaging},
  year    = {2013},
  volume  = {4},
  number  = {6},
  pages   = {741--751},
  doi     = {10.1007/s13244-013-0275-3}
}

@article{tomography_US_variability,
  author  = {Mangona, Kate Louise M. and
             Guillerman, R. Paul and
             Mangona, Victor S. and
             Carpenter, Jennifer and
             Zhang, Wei and
             Lopez, Monica and
             Orth, Robert C.},
  title   = {Diagnostic Performance of Ultrasonography for Pediatric Appendicitis: A Night and Day Difference?},
  journal = {Academic Radiology},
  year    = {2017},
  volume  = {24},
  number  = {12},
  pages   = {1616--1620},
  doi     = {10.1016/j.acra.2017.06.007}
}

@article{ct_peds_risk,
  author  = {Shah, Sohail R. and
             Sinclair, Kelly A. and
             Theut, Stephanie B. and
             Johnson, Kathy M. and
             Holcomb, III, George W. and
             St Peter, Shawn D.},
  title   = {Computed Tomography Utilization for the Diagnosis of Acute Appendicitis in Children Decreases With a Diagnostic Algorithm},
  journal = {Annals of Surgery},
  year    = {2016},
  volume  = {264},
  number  = {3},
  pages   = {474--481},
  doi     = {10.1097/SLA.0000000000001867}
}

@article{regensburg_ml_appendicitis,
  author  = {Marcinkevi{\v{c}}s, Ri{\v{c}}ards and Reis Wolfertstetter, Patricia and Wellmann, Sven and Knorr, Christian and Vogt, Julia E.},
  title   = {Using Machine Learning to Predict the Diagnosis, Management and Severity of Pediatric Appendicitis},
  journal = {Frontiers in Pediatrics},
  year    = {2021},
  volume  = {9},
  pages   = {662183},
  doi     = {10.3389/fped.2021.662183},
  url     = {https://www.frontiersin.org/articles/10.3389/fped.2021.662183/full}
}

@article{ml_appendicitis_wjes,
  author  = {Schipper, Anoeska and
             Belgers, Peter and
             O'Connor, Rory and
             Jie, Kim Ellis and
             Dooijes, Robin and
             Bosma, Joeran Sander and
             Kurstjens, Steef and
             Kusters, Ron and
             van Ginneken, Bram and
             Rutten, Matthieu},
  title   = {Machine-Learning Based Prediction of Appendicitis for Patients Presenting with Acute Abdominal Pain at the Emergency Department},
  journal = {World Journal of Emergency Surgery},
  year    = {2024},
  volume  = {19},
  pages   = {40},
  doi     = {10.1186/s13017-024-00570-7},
  url     = {https://link.springer.com/article/10.1186/s13017-024-00570-7}
}

@article{finlayson2021clinician,
  author  = {Finlayson, Samuel G. and
             Subbaswamy, Adarsh and
             Singh, Karandeep and
             Bowers, John and
             Kupke, Annabel and
             Zittrain, Jonathan and
             Kohane, Isaac S. and
             Saria, Suchi},
  title   = {The Clinician and Dataset Shift in Artificial Intelligence},
  journal = {New England Journal of Medicine},
  year    = {2021},
  volume  = {385},
  number  = {3},
  pages   = {283--286},
  doi     = {10.1056/NEJMc2104626}
}

@article{liu2024chatgpt,
  author  = {Liu, M. and Okuhara, T. and Chang, X. and Shirabe, R. and Nishiie, Y. and Okada, H. and Kiuchi, T.},
  title   = {Performance of {ChatGPT} Across Different Versions in Medical Licensing Examinations Worldwide: Systematic Review and Meta-Analysis},
  journal = {Journal of Medical Internet Research},
  year    = {2024},
  volume  = {26},
  pages   = {e60807},
  doi     = {10.2196/60807},
  url     = {https://www.jmir.org/2024/1/e60807/}
}

@article{brin2024gpt,
  author  = {Brin, Dana and
             Sorin, Vera and
             Konen, Eli and
             Nadkarni, Girish and
             Glicksberg, Benjamin S. and
             Klang, Eyal},
  title   = {How {GPT} Models Perform on the United States Medical Licensing Examination: A Systematic Review},
  journal = {Discover Applied Sciences},
  year    = {2024},
  volume  = {6},
  number  = {10},
  pages   = {500},
  doi     = {10.1007/s42452-024-06194-5}
}

@article{lm_medicine,
  author  = {Thirunavukarasu, Arun James and
             Ting, Darren Shu Jeng and
             Elangovan, Kabilan and
             Gutierrez, Laura and
             Tan, Ting Fang and
             Ting, Daniel Shu Wei},
  title   = {Large language models in medicine},
  journal = {Nature Medicine},
  year    = {2023},
  volume  = {29},
  number  = {8},
  pages   = {1930--1940},
  doi     = {10.1038/s41591-023-02448-8}
}

@article{lee2023benefits,
  author  = {Lee, Peter and Bubeck, Sebastien and Petro, Joseph},
  title   = {Benefits, Limits, and Risks of {GPT-4} as an {AI} Chatbot for Medicine},
  journal = {New England Journal of Medicine},
  year    = {2023},
  volume  = {388},
  number  = {13},
  pages   = {1233--1239},
  doi     = {10.1056/NEJMsr2214184},
  url     = {https://doi.org/10.1056/NEJMsr2214184}
}

@misc{fda2021gmlp,
  author = {{U.S. Food and Drug Administration} and {Health Canada} and {Medicines and Healthcare products Regulatory Agency}},
  title  = {Good Machine Learning Practice for Medical Device Development: Guiding Principles},
  year   = {2021},
  note   = {Accessed June 14, 2026},
  url    = {https://www.fda.gov/medical-devices/software-medical-device-samd/good-machine-learning-practice-medical-device-development-guiding-principles}
}

@article{kalyan2020secnlp,
  author  = {Kalyan, Katikapalli Subramanyam and
             Sangeetha, Sivanesan},
  title   = {SECNLP: A Survey of Embeddings in Clinical Natural Language Processing},
  journal = {Journal of Biomedical Informatics},
  year    = {2020},
  volume  = {101},
  pages   = {103323},
  doi     = {10.1016/j.jbi.2019.103323},
  note    = {Also available as arXiv:1903.01039}
}

@article{odonnell2009copy,
  author  = {{O'Donnell}, Heather C. and Kaushal, Rainu and Barr{\'o}n, Yolanda and Callahan, Mark A. and Adelman, Ronald D. and Siegler, Eugenia L.},
  title   = {Physicians' Attitudes Towards Copy and Pasting in Electronic Note Writing},
  journal = {Journal of General Internal Medicine},
  year    = {2009},
  volume  = {24},
  pages   = {63--68},
  doi     = {10.1007/s11606-008-0843-2},
  note    = {Published online November 8, 2008}
}

@article{hirschtick2006copy,
  author  = {Hirschtick, Robert E.},
  title   = {Copy-and-Paste},
  journal = {JAMA},
  year    = {2006},
  volume  = {295},
  number  = {20},
  pages   = {2335--2336},
  doi     = {10.1001/jama.295.20.2335}
}

@article{neha2026radiomics,
  author        = {Neha, Fnu and Shukla, Deepak Kumar},
  title         = {Radiomics in Medical Imaging: Methods, Applications, and Challenges},
  journal       = {arXiv preprint arXiv:2602.00102},
  year          = {2026},
  eprint        = {2602.00102},
  archivePrefix = {arXiv},
  url           = {https://arxiv.org/abs/2602.00102}
}

@article{alvarado1986practical,
  author  = {Alvarado, Alfredo},
  title   = {A Practical Score for the Early Diagnosis of Acute Appendicitis},
  journal = {Annals of Emergency Medicine},
  year    = {1986},
  volume  = {15},
  number  = {5},
  pages   = {557--564},
  doi     = {10.1016/S0196-0644(86)80993-3}
}

@article{samuel2002pediatric,
  author  = {Samuel, Madan},
  title   = {Pediatric Appendicitis Score},
  journal = {Journal of Pediatric Surgery},
  year    = {2002},
  month   = jun,
  volume  = {37},
  number  = {6},
  pages   = {877--881},
  doi     = {10.1053/jpsu.2002.32893}
}

@article{di_saverio2020_wses_appendicitis,
  author  = {Salomone Di Saverio and Mauro Podda and Belinda De Simone and Marco Ceresoli and Goran Augustin and Alice Gori and Marja Boermeester and Massimo Sartelli and Federico Coccolini and Antonio Tarasconi and Nicola de’ Angelis and Dieter G. Weber and Matti Tolonen and Arianna Birindelli and Walter Biffl and Ernest E. Moore and Michael Kelly and Kjetil Soreide and Jeffry Kashuk and Richard Ten Broek and Carlos Augusto Gomes and Michael Sugrue and Richard Justin Davies and Dimitrios Damaskos and Ari Leppäniemi and Andrew Kirkpatrick and Andrew B. Peitzman and Gustavo P. Fraga and Ronald V. Maier and Raul Coimbra and Massimo Chiarugi and Gabriele Sganga and Adolfo Pisanu and Gian Luigi de’ Angelis and Edward Tan and Harry Van Goor and Francesco Pata and Isidoro Di Carlo and Osvaldo Chiara and Andrey Litvin and Fabio C. Campanile and Boris Sakakushev and Gia Tomadze and Zaza Demetrashvili and Rifat Latifi and Fakri Abu-Zidan and Oreste Romeo and Helmut Segovia-Lohse and Gianluca Baiocchi and David Costa and Sandro Rizoli and Zsolt J. Balogh and Cino Bendinelli and Thomas Scalea and Rao Ivatury and George Velmahos and Roland Andersson and Yoram Kluger and Luca Ansaloni and Fausto Catena},
  title   = {Diagnosis and Treatment of Acute Appendicitis: 2020 Update of the {WSES} Jerusalem Guidelines},
  journal = {World Journal of Emergency Surgery},
  year    = {2020},
  month   = apr,
  volume  = {15},
  number  = {1},
  pages   = {27},
  doi     = {10.1186/s13017-020-00306-3}
}

@article{lam2023ai_appendicitis_review,
  author  = {Lam, Antoinette and Squires, Emily and Tan, Sheryn and Swen, Ng Jeng and Barilla, Adriano and Kovoor, Joshua and Gupta, Aashray and Bacchi, Stephen and Khurana, Sanjeev},
  title   = {Artificial Intelligence for Predicting Acute Appendicitis: A Systematic Review},
  journal = {ANZ Journal of Surgery},
  year    = {2023},
  volume  = {93},
  number  = {9},
  pages   = {2070--2078},
  doi     = {10.1111/ans.18610},
  pmid    = {37458222},
  url     = {https://pubmed.ncbi.nlm.nih.gov/37458222/}
}

@article{nori2023capabilities,
  author        = {Nori, Harsha and King, Nicholas and McKinney, Scott Mayer and Carignan, Dean and Horvitz, Eric},
  title         = {Capabilities of {GPT-4} on Medical Challenge Problems},
  journal       = {arXiv preprint arXiv:2303.13375},
  year          = {2023},
  eprint        = {2303.13375},
  archivePrefix = {arXiv},
  primaryClass  = {cs.CL},
  doi           = {10.48550/arXiv.2303.13375}
}

@article{singhal_medpalm2,
  author  = {Singhal, Karan and Tu, Tao and Gottweis, Juraj and Sayres, Rory and Wulczyn, Ellery and Amin, Mohamed and Hou, Le and Clark, Kevin and Pfohl, Stephen R. and Cole-Lewis, Heather and Neal, Darlene and Rashid, Qazi Mamunur and Schaekermann, Mike and Wang, Amy and Dash, Dev and Chen, Jonathan H. and Shah, Nigam H. and Lachgar, Sami and Mansfield, Philip Andrew and Prakash, Sushant and Green, Bradley and Dominowska, Ewa and Ag{\"u}era y Arcas, Blaise and Toma{\v{s}}ev, Nenad and Liu, Yun and Wong, Renee and Semturs, Christopher and Mahdavi, S. Sara and Barral, Joelle K. and Webster, Dale R. and Corrado, Greg S. and Matias, Yossi and Azizi, Shekoofeh and Karthikesalingam, Alan and Natarajan, Vivek},
  title   = {Toward expert-level medical question answering with large language models},
  journal = {Nature Medicine},
  year    = {2025},
  volume  = {31},
  number  = {3},
  pages   = {943--950},
  doi     = {10.1038/s41591-024-03423-7}
}

@article{hager_llm_clinical,
  author  = {Hager, Paul and Jungmann, Friederike and Holland, Robbie and Bhagat, Kunal and Hubrecht, Inga and Knauer, Manuel and Vielhauer, Jakob and Makowski, Marcus and Braren, Rickmer and Kaissis, Georgios and Rueckert, Daniel},
  title   = {Evaluation and Mitigation of the Limitations of Large Language Models in Clinical Decision-Making},
  journal = {Nature Medicine},
  year    = {2024},
  volume  = {30},
  number  = {9},
  pages   = {2613--2622},
  month   = sep,
  doi     = {10.1038/s41591-024-03097-1},
  url     = {https://doi.org/10.1038/s41591-024-03097-1}
}

@article{kim_llm_reasoning,
  author  = {Goh, Ethan and Gallo, Robert and Hom, Jason and Strong, Eric and Weng, Yingjie and Kerman, Hannah and Cool, Joséphine A. and Kanjee, Zahir and Parsons, Andrew S. and Ahuja, Neera and Horvitz, Eric and Yang, Daniel and Milstein, Arnold and Olson, Andrew P. J. and Rodman, Adam and Chen, Jonathan H.},
  title   = {Large Language Model Influence on Diagnostic Reasoning: A Randomized Clinical Trial},
  journal = {JAMA Network Open},
  year    = {2024},
  volume  = {7},
  number  = {10},
  pages   = {e2440969},
  doi     = {10.1001/jamanetworkopen.2024.40969},
  url     = {https://jamanetwork.com/journals/jamanetworkopen/fullarticle/2825395}
}

@article{clinical_summarization_stanford,
  author  = {Van Veen, Dave and Van Uden, Cara and Blankemeier, Louis and Delbrouck, Jean-Benoit and Aali, Asad and Bluethgen, Christian and Pareek, Anuj and Polacin, Malgorzata and Reis, Eduardo Pontes and Seehofnerová, Anna and Rohatgi, Nidhi and Hosamani, Poonam and Collins, William and Ahuja, Neera and Langlotz, Curtis P. and Hom, Jason and Gatidis, Sergios and Pauly, John and Chaudhari, Akshay S.},
  title   = {Adapted large language models can outperform medical experts in clinical text summarization},
  journal = {Nature Medicine},
  year    = {2024},
  volume  = {30},
  number  = {4},
  pages   = {1134--1142},
  doi     = {10.1038/s41591-024-02855-5}
}

@article{Neha2025_RAG_Healthcare,
  author  = {Neha, Fnu and Bhati, Deepshikha and Shukla, Deepak Kumar},
  title   = {Retrieval-Augmented Generation ({RAG}) in Healthcare: A Comprehensive Review},
  journal = {AI},
  year    = {2025},
  volume  = {6},
  number  = {9},
  pages   = {226},
  doi     = {10.3390/ai6090226}
}

@inproceedings{chen2023rgb,
  author    = {Chen, Jiawei and Lin, Hongyu and Han, Xianpei and Sun, Le},
  title     = {Benchmarking Large Language Models in Retrieval-Augmented Generation},
  booktitle = {Proceedings of the Thirty-Eighth AAAI Conference on Artificial Intelligence},
  year      = {2024},
  pages     = {17754--17762},
  doi       = {10.1609/aaai.v38i16.29728}
}

@inproceedings{hu2024prompt_perturbation_rag,
  author    = {Hu, Zhibo and Wang, Chen and Shu, Yanfeng and Paik, Hye-Young and Zhu, Liming},
  title     = {Prompt Perturbation in Retrieval-Augmented Generation Based Large Language Models},
  booktitle = {Proceedings of the 30th ACM SIGKDD Conference on Knowledge Discovery and Data Mining},
  year      = {2024},
  pages     = {1119--1130},
  doi       = {10.1145/3637528.3671932}
}

@article{external_validation,
  author  = {Marcinkevi{\v{c}}s, Ri{\v{c}}ards and
             Sokol, Kacper and
             Paulraj, Akhil and
             Hilbert, Melinda A. and
             Rimili, Vivien and
             Wellmann, Sven and
             Knorr, Christian and
             Reingruber, Bertram and
             Vogt, Julia E. and
             Reis Wolfertstetter, Patricia},
  title   = {{External Validation of Predictive Models for Diagnosis, Management and Severity of Pediatric Appendicitis}},
  journal = {Frontiers in Pediatrics},
  year    = {2025},
  volume  = {13},
  pages   = {1587488},
  doi     = {10.3389/fped.2025.1587488},
  url     = {https://www.frontiersin.org/articles/10.3389/fped.2025.1587488/full}
}

@article{scott2013_datatotext,
  author  = {Scott, Donia and Hallett, Catalina and Fettiplace, Rachel},
  title   = {Data-to-Text Summarisation of Patient Records: Using Computer-Generated Summaries to Access Patient Histories},
  journal = {Patient Education and Counseling},
  year    = {2013},
  volume  = {92},
  number  = {2},
  pages   = {153--159},
  doi     = {10.1016/j.pec.2013.04.019}
}

@article{lee2018_nlg_ehr,
  author  = {Lee, Scott H.},
  title   = {Natural Language Generation for Electronic Health Records},
  journal = {npj Digital Medicine},
  year    = {2018},
  volume  = {1},
  pages   = {63},
  doi     = {10.1038/s41746-018-0070-0}
}

@article{lee2025_pseudonotes,
  author  = {Lee, Simon A. and Jain, Sujay and Chen, Alex and
             Ono, Kyoka and Biswas, Arabdha and
             Rudas, {Ákos} and Fang, Jennifer and
             Chiang, Jeffrey N.},
  title   = {Clinical Decision Support Using Pseudo-Notes from Multiple Streams of {EHR} Data},
  journal = {npj Digital Medicine},
  year    = {2025},
  volume  = {8},
  pages   = {394},
  doi     = {10.1038/s41746-025-01777-x}
}

@article{rabaey2025simsum,
  author  = {Rabaey, Paloma and Heytens, Stefan and Demeester, Thomas},
  title   = {{SimSUM} -- simulated benchmark with structured and unstructured medical records},
  journal = {Journal of Biomedical Semantics},
  year    = {2025},
  volume  = {16},
  number  = {1},
  pages   = {20},
  doi     = {10.1186/s13326-025-00341-6}
}

@article{sellergren2025_medgemma,
  author = {
    Andrew Sellergren and
    Sahar Kazemzadeh and
    Tiam Jaroensri and
    Atilla Kiraly and
    Madeleine Traverse and
    Timo Kohlberger and
    Shawn Xu and
    Fayaz Jamil and
    C{\'i}an Hughes and
    Charles Lau and
    Justin Chen and
    Fereshteh Mahvar and
    Liron Yatziv and
    Tiffany Chen and
    Bram Sterling and
    Stefanie Anna Baby and
    Susanna Maria Baby and
    Jeremy Lai and
    Samuel Schmidgall and
    Lu Yang and
    Kejia Chen and
    Per Bjornsson and
    Shashir Reddy and
    Ryan Brush and
    Kenneth Philbrick and
    Mercy Asiedu and
    Ines Mezerreg and
    Howard Hu and
    Howard Yang and
    Richa Tiwari and
    Sunny Jansen and
    Preeti Singh and
    Yun Liu and
    Shekoofeh Azizi and
    Aishwarya Kamath and
    Johan Ferret and
    Shreya Pathak and
    Nino Vieillard and
    Ramona Merhej and
    Sarah Perrin and
    Tatiana Matejovicova and
    Alexandre Ram{\'e} and
    Morgane Riviere and
    Louis Rouillard and
    Thomas Mesnard and
    Geoffrey Cideron and
    Jean-Bastien Grill and
    Sabela Ramos and
    Edouard Yvinec and
    Michelle Casbon and
    Elena Buchatskaya and
    Jean-Baptiste Alayrac and
    Dmitry Lepikhin and
    Vlad Feinberg and
    Sebastian Borgeaud and
    Alek Andreev and
    Cassidy Hardin and
    Robert Dadashi and
    L{\'e}onard Hussenot and
    Armand Joulin and
    Olivier Bachem and
    Yossi Matias and
    Katherine Chou and
    Avinatan Hassidim and
    Kavi Goel and
    Clement Farabet and
    Joelle Barral and
    Tris Warkentin and
    Jonathon Shlens and
    David Fleet and
    Victor Cotruta and
    Omar Sanseviero and
    Gus Martins and
    Phoebe Kirk and
    Anand Rao and
    Shravya Shetty and
    David F. Steiner and
    Can Kirmizibayrak and
    Rory Pilgrim and
    Daniel Golden and
    Lin Yang
  },
  title         = {MedGemma Technical Report},
  journal       = {arXiv preprint arXiv:2507.05201},
  year          = {2025},
  doi           = {10.48550/arXiv.2507.05201},
  eprint        = {2507.05201},
  archivePrefix = {arXiv}
}

@article{grattafiori2024_llama3,
 author        = {Grattafiori, Aaron and Dubey, Abhimanyu and Jauhri, Abhinav and Pandey, Abhinav and Kadian, Abhishek and Al-Dahle, Ahmad and Letman, Aiesha and Mathur, Akhil and Schelten, Alan and Vaughan, Alex and others},
  title         = {The {Llama 3} Herd of Models},
  journal       = {arXiv preprint arXiv:2407.21783},
  year          = {2024},
  doi           = {10.48550/arXiv.2407.21783},
  eprint        = {2407.21783},
  archivePrefix = {arXiv}
}

@techreport{openai2026_gpt55_system_card,
  author      = {{OpenAI}},
  title       = {{GPT-5.5} System Card},
  institution = {OpenAI},
  year        = {2026},
  month       = apr,
  url         = {https://openai.com/index/gpt-5-5-system-card/},
  note        = {Accessed June 12, 2026}
}

@techreport{openai2025_o3_o4mini_system_card,
  author      = {{OpenAI}},
  title       = {{OpenAI o3 and o4-mini} System Card},
  institution = {OpenAI},
  year        = {2025},
  url         = {https://cdn.openai.com/pdf/2221c875-02dc-4789-800b-e7758f3722c1/o3-and-o4-mini-system-card.pdf},
  note        = {Accessed June 12, 2026}
}

@techreport{anthropic2026_claude_opus47_system_card,
  author      = {{Anthropic}},
  title       = {{Claude Opus 4.7} System Card},
  institution = {Anthropic},
  year        = {2026},
  month       = apr,
  url         = {https://anthropic.com/claude-opus-4-7-system-card},
  note        = {Published April 16, 2026; accessed June 12, 2026}
}

@article{wornow2023shaky,
  author  = {Wornow, Michael and
             Xu, Yizhe and
             Thapa, Rahul and
             Patel, Birju and
             Steinberg, Ethan and
             Fleming, Scott L. and
             Pfeffer, Michael A. and
             Fries, Jason and
             Shah, Nigam H.},
  title   = {The Shaky Foundations of Large Language Models and Foundation Models for Electronic Health Records},
  journal = {npj Digital Medicine},
  year    = {2023},
  volume  = {6},
  number  = {1},
  pages   = {135},
  doi     = {10.1038/s41746-023-00879-8}
}

@article{maity_llm_healthcare_review,
  author  = {Maity, Subhankar and Saikia, Manob Jyoti},
  title   = {Large Language Models in Healthcare and Medical Applications: A Review},
  journal = {Bioengineering},
  year    = {2025},
  volume  = {12},
  number  = {6},
  pages   = {631},
  doi     = {10.3390/bioengineering12060631},
  url     = {https://www.mdpi.com/2306-5354/12/6/631}
}

@article{marcinkevics2024interpretable,
  author  = {Marcinkevi{\v{c}}s, Ri{\v{c}}ards and Reis Wolfertstetter, Patricia and Klimiene, Ugne and Chin-Cheong, Kieran and Paschke, Alyssia and Zerres, Julia and Denzinger, Markus and Niederberger, David and Wellmann, Sven and Ozkan, Ece and Knorr, Christian and Vogt, Julia E.},
  title   = {Interpretable and intervenable ultrasonography-based machine learning models for pediatric appendicitis},
  journal = {Medical Image Analysis},
  year    = {2024},
  volume  = {91},
  pages   = {103042},
  doi     = {10.1016/j.media.2023.103042}
}

@article{kopf2023openassistant,
  title={Openassistant conversations-democratizing large language model alignment},
  author={K{\"o}pf, Andreas and Kilcher, Yannic and Von R{\"u}tte, Dimitri and Anagnostidis, Sotiris and Tam, Zhi Rui and Stevens, Keith and Barhoum, Abdullah and Nguyen, Duc and Stanley, Oliver and Nagyfi, Rich{\'a}rd and others},
  journal={Advances in neural information processing systems},
  volume={36},
  pages={47669--47681},
  year={2023}
}

@inproceedings{chen2016xgboost,
	title={Xgboost: A scalable tree boosting system},
	author={Chen, Tianqi and Guestrin, Carlos},
	booktitle={Proceedings of the 22nd acm sigkdd international conference on knowledge discovery and data mining},
	pages={785--794},
	year={2016}
}

\end{document}